\documentclass{article} 
\usepackage{iclr2026_conference,times}

\iclrfinalcopy

\usepackage{listings}
\usepackage{adjustbox}
\usepackage{booktabs}
\usepackage{enumitem}
\RequirePackage{tgtermes}
\RequirePackage{newtxtext}
\RequirePackage{newtxmath}
\RequirePackage{bm}
\RequirePackage{endnotes}

\usepackage{multirow}
\usepackage{makecell}
\usepackage{array}
\usepackage{hyperref}
\usepackage{subcaption}
\usepackage{tcolorbox}
\tcbuselibrary{skins}

\usepackage{anyfontsize}
\usepackage[T1]{fontenc}
\usepackage{algorithm}
\usepackage{algpseudocode}
\usepackage{tikz}
\usetikzlibrary{
  arrows.meta,
  positioning,
  shapes,
  calc,
  fit,
  backgrounds
}

\newcommand{\best}[1]{\textbf{#1}}

\usepackage{natbib}
 \bibpunct[, ]{(}{)}{,}{a}{}{,}%

\title{\textsc{Text2Model}: Modeling Copilots for Text-to-Model Translation}

\author{Serdar Kad{\i}o\u{g}lu\textsuperscript{1, 2}, Karthik Uppuluri\textsuperscript{1}, and Akash Singirikonda\textsuperscript{2}\\
\textsuperscript{1}AI Center of Excellence, Fidelity Investments \\
\textsuperscript{2}Department of Computer Science, Brown University\\
\texttt{\{serdark@cs.brown.edu\}}
}

\begin{document}

\maketitle
\begin{abstract}
There is growing interest in leveraging large language models (LLMs) for text-to-model translation. This paper aims to advance this line of research by introducing \textsc{Text2Model} and \textsc{Text2Zinc}. \textsc{Text2Model} is a suite of copilots based on several LLM strategies with varying complexity, along with an online leaderboard. \textsc{Text2Zinc} is a cross-domain dataset for capturing optimization and satisfaction problems specified in natural language, along with an interactive editor with built-in AI assistant. While there is an emerging literature on using LLMs for translating combinatorial problems into formal models, our work is the first attempt to  integrate \textit{both} satisfaction and optimization problems within a \textit{unified architecture} and \textit{dataset}. Moreover, our approach is  \textit{solver-agnostic} unlike existing work that focuses on translation to a solver-specific model. To achieve this, we leverage \textsc{MiniZinc}'s solver-and-paradigm-agnostic modeling capabilities to formulate combinatorial problems. We conduct comprehensive experiments to compare execution and solution accuracy across several single- and multi-call strategies, including; zero-shot prompting, chain-of-thought reasoning, intermediate representations via knowledge-graphs, grammar-based syntax encoding, and agentic approaches that decompose the model into sequential sub-tasks. Our copilot strategies are competitive, and in parts improve, recent research in this domain. Our findings indicate that while LLMs are promising they are not yet a push-button technology for combinatorial modeling. We contribute \textsc{Text2Model} copilots and leaderboard, and \textsc{Text2Zinc} and interactive editor to open-source to support closing this performance gap. 
\end{abstract}


\section{Introduction}
\label{sec:intro}
Optimization Technology has achieved significant advancements, ranging from dramatic improvements in solver efficiency to the development of high-level modelling languages. Nevertheless, the fundamental decision-making framework has remained unchanged for decades, adhering to the de facto \textit{model-and-run strategy}. Within this status quo, users are required to manually convert problem descriptions into formal models, which are subsequently processed by solvers to obtain solutions.




Over the years, high-level modeling languages such as \textsc{MiniZinc}~\citep{minizinc}, CPMpy~\citep{guns2019increasing}, and GAMS~\citep{Bussieck2004} have partially addressed the accessibility challenge by providing solver-agnostic approaches that are powerful and flexible. These modelling frameworks enable practitioners to focus on describing their problems without worrying about specific solution methods, making them especially useful for real-world applications, where requirements often change over time. However, the cognitive barrier of translating problem descriptions into formal constraint models persists. This barrier is particularly acute, as domain experts who deeply understand their problem domain often lack the specialized knowledge required for formal modeling. The resulting dependency on modeling experts creates operational bottlenecks and can lead to misinterpretation of domain-specific requirements during the translation process.


In parallel, Large-Language Models (LLMs) have emerged as the communication medium with machines~\citep{openai2024gpt4technicalreport,geminiteam2025geminifamilyhighlycapable,deepseekai2025deepseekv3technicalreport}. While language models are powerful at interfacing with natural text, they struggle with the consistency and precision required in formal, declarative approaches, from basic type declarations to complex constraint relations. They face significant challenges in handling the mathematical and logical reasoning needed for automated modeling~\citep{SimchiLevi2025Democratizing,Wasserkrug2025,ner4opt2024}. This gap between understanding textual descriptions and turning them into problem formulations indicates that more work is needed for modeling assistants. This is precisely the topic of this paper. 

The ultimate vision motivating this paper is a paradigm shift by integrating automated modelling assistants capable of translating problem descriptions expressed in natural language into formal optimization formulations. We refer to this as Modeling copilots analogous to Coding copilots. With this goal in mind, in this paper, we make the following contributions: 



\vspace{-0.2cm}
\subsection{Our Contributions}
\vspace{-0.2cm}
\label{sec:Contibutions}

Our main contributions are twofolds: i) \textsc{Text2Model}
problem formulation and several copilots with varying levels of complexity, along with an online leaderboard for the community to test different strategies, and ii) \textsc{Text2Zinc}
dataset to enable the community to benchmark text-to-model translation problem, along with an interactive editor for ongoing data curation.

 Our first contribution, \textsc{Text2Model}, is the formalization and design of several LLM strategies for building constraint models from problem descriptions. Beyond zero-shot prompting as a baseline, we employ Chain-of-Thought, Knowledge Graphs, Grammar Encodings, and Agentic Frameworks.
 
 The use of knowledge graph representation of the underlying constraint network and the grammar encoding of the constraint modeling language for code generation are novel contributions of our work that have not been studied before. 
 
 We open-source the \textsc{Text2Model} collection\footnote{\url{https://skadio.github.io/text2model/}}, a suite of LLM modeling copilots, datasets, fined-tuned models, demos, interactive editor, and online leaderboard for translating natural language text into formal combinatorial constraint models. 
 We present numerical results to assess the execution accuracy (does the generated model compile and run?) and the solution accuracy (is the solution found correct/optimal?). These results form the baseline of our online \textsc{Text2Model} leaderboard, which is ready to accommodate new methods (or new LLMs) as they arise.
 
Our second contribution, \textsc{Text2Zinc}, is a unified and cross-domain dataset that combines \textit{both satisfaction and optimization problems} expressed in natural language. This is the \textit{first dataset} in this line of research that enables modeling both satisfaction and optimization problems. Existing datasets only cover one objective; either feasibility or optimization. In addition, our modelling approach remains \textit{agnostic to the underlying solving technology} addressing a significant gap in literature focus only on a specific solver. In our approach, the generated models can be compiled into different backends, thanks to our \textsc{MiniZinc} integration, including Constraint Programming solvers, such as Gecode and Chuffed, SAT/LCG solvers, such as Google Or-Tools, and Mathematical Programming solvers, SCIP and Gurobi. 
  
  The \textsc{Text2Zinc} dataset contains a diverse range of problem types across multiple domains, carefully curated from excellent resources including \textsc{LPW}
  ~\citep{lpwp}, \textsc{Nlp4lp}
~\citep{optimus}, \textsc{ComplexOR}
~\citep{complexor}, \textsc{CspLib}
\textsc{Hakank's} collection
~\citep{hakank}, \textsc{IndustryOR}, \textsc{MAMO}, \textsc{NL4Opt}~\citep{Huang_2025}. With the emergence of different datasets with varying format and schema, it is worth studying how to design text-to-model benchmarks. We show \textit{how to unify} these different sources, problem types, and resources using a generic schema designed to work with LLMs. Further, we enhance these problems through reformulation, metadata enrichment, curation, and manual verification to ensure consistency and quality. This comprehensive collection provides a robust foundation for evaluating text-to-model translation. To support data curation going forward, we also release the \textsc{Text2Zinc Editor} to add and update new instances.

Finally, our copilot pipeline, online leaderboard, dataset, and interactive editor are all contributed to open-source to support further research at the intersection of LLMs and Optimization. 

The remainder of this paper is organized as follows. The background (\S \hyperref[sec:background]{2}) reviews fundamentals of combinatorial problems, brief overview of \textsc{MiniZinc}, chain-of-thought, knowledge graphs, and grammar encodings in LLMs.  We then formalize the text-to-model translation problem (\S\hyperref[sec:formulation]{3}) and present our different \textsc{Text2Model} copilots (\S\hyperref[sec:formulation]{4}). In \S\hyperref[sec:dataset]{5}, we present the details of \textsc{Text2Zinc} dataset. In \S\hyperref[sec:results]{6} we present computational results and discuss generation observations. 

\section{Background}
\label{sec:background}
This paper is built on three main areas; combinatorial problems, constraint modeling, and intermediary representations in between unstructured text (i.e., problem description) and structured code (i.e., model formulation). To provide background on these, let us briefly review satisfaction and optimization problems, the \textsc{MiniZinc} modeling language, and LLMs. Readers familiar with these can jump to \textsc{Text2Model} problem definition in the next section (\S \hyperref[sec:formulation]{3}).

\subsection{Constraint Satisfaction and Optimization}

Constraint Satisfaction Problem (CSPs) can be defined as a triple across variables, domains and constraints: 
$$CSP = (X, D, C)$$
\begin{itemize}
  \item $X = \{X_1, X_2, \dots, X_n\}$ is a finite set of decision variables;
  \item $D = \{D_{1}, D_{2}, \dots, D_{n}\}$ are the domains of each variable, which assign each variable $X_i$ to a non-empty domain $D_{i}$ of admissible values;
  \item $C = \{c_1, c_2, \dots, c_m\}$ is the set of constraints, each which maps a total assignment of variables to values in their domains to a truth value.
\end{itemize}

A solution to a CSP is an assignment of all variables $x \in D_{X_1} \times\cdots\times D_{X_n}$ such that all constraints in $C$ evaluate to \textit{true}, i.e., satisfied.

A natural extension to constraint satisfaction problems are Constraint Optimization Problems (COPs), which augments the definition of a CSP with an objective function:
$$ COP = (X, D, C, O)$$
where $O : D_{X_1}\times\cdots\times D_{X_n} \to \mathbb{R}$ assigns a cost (or value) to every complete assignment.  A feasible assignment that minimizes (or maximizes) $O$ while respecting all constraints is called an \emph{optimal solution}.

A wealth of literature exists to solve CSPs and COPs, and of particular interest to our paper are declarative paradigms such as Constraint Programming (CP), typically used for solving satisfaction problems, and Mathematical Programming (MP), typically used for solving optimization problems~\citep{CPandOR}. 

In the classical Operations Research (OR) pipeline, a modeling expert specifies the tuples $(X,D,C)$ for CSPs or $(X,D,C,O)$ for COPs, in a high-level declarative modeling language and then passes this specification, together with optional search and other configuration parameter settings, to an off-the-shelf \textit{constraint solver}. Our focus is on assisting the modeler infer these tuples automatically from problem statements. To that end, \textsc{Text2Zinc} pairs textual problem description with (i) a \textsc{MiniZinc} model that captures the corresponding CSP or COP, and (ii), reference solutions or checkers that certify correctness. 

Let us now review \textsc{MiniZinc}.

\begin{lstlisting}[numbers=none, captionpos=b, basicstyle=\normalsize\ttfamily, caption={All different constraint in MiniZinc.}]

% pairwise binary inequalities 
predicate all_different(array[int] of var float:x) =
forall(i,j in index_set(x) where i < j)(x[i] != x[j]); 

% built-in global constraint
predicate all_different(array[int] of var int:x) =
gecode_all_different(x); % native Gecode version
\end{lstlisting}

\subsection{\textsc{MiniZinc} Modelling Language}
In this work, we model both CSPs and COPs in \textsc{MiniZinc}\footnote{\url{https://www.minizinc.org/}}, a high-level constraint modeling language that supports both discrete and continuous optimization and satisfaction problems~\citep{minizinc}. \textsc{MiniZinc}'s solver-agnostic design allows communication with various solver backends. This enables leveraging different problem-solving paradigms, including Constraint Programming (CP), (Mixed) Integer Programming (MIP), and Boolean Satisfiability and Lazy Clause Generation (SAT/LCG). This flexibility is achieved through compilation to \textsc{FlatZinc}, an intermediate language that interfaces with different solvers, allowing the same \textsc{MiniZinc} model to be used across multiple backends without any code modifications. This allows end users to write the modeling code once and test different backends that best suit the task at hand. 

A key feature of \textsc{MiniZinc} is its use of global constraints, a concept originating in CP, which significantly simplifies the modeling process. For example, the \textit{all\_different($x_1, .. x_n$)} constraint specifies that variables $x_1, x_2, ... x_n$ must take distinct values, replacing numerous pairwise inequality constraints that would be required in traditional MIP solvers. These global constraints can be further specialized for particular solvers, often leading to improved performance~\cite{globalconstraints,minizinc}. To make this and \textsc{MiniZinc} language more concrete, Listing 1 shows an example the default expansion of the \textit{all\_different }constraint from \textsc{MiniZinc} to \textsc{FlatZinc} as binary inequalities. For the same predicate, if the solver is interfacing with \textsc{Gecode}, it can use \textsc{Gecode}'s native all\_different constraint instead. This way, global constraints allow users to leverage higher-level abstractions rather than focusing on low-level decomposition.

\textsc{MiniZinc}'s practical utility for real-world applications is further enhanced by its clear separation of \textbf{models} (\textbf{.mzn }files) and \textbf{instances} (\textbf{.dzn} files). Models contain the problem structure, while instances provide specific input data, allowing a single model for the problem to be reused across multiple instances of the same  problem. The language structure is straightforward, consisting of four main components: decision variables, constraints, parameters, a satisfaction goal for feasibility, or an objective function for optimization problems. This structure fosters creating modular programs. Additionally, \textsc{MiniZinc} supports automated solution checking and model validation, which help evaluate model correctness during development.

Given its advanced features, we consider \textsc{MiniZinc} as an excellent fit for bridging natural language and solver formulations. We hypothesize that higher-level language constructs, as offered by \textsc{MiniZinc}, serve as a good interface between LLMs and combinatorial formulations. Let us now go over Chain-of-Thought, Knowledge Graphs, and Grammar Encodings.

\subsection{Chain-of-Thought}
Chain-of-Thought (CoT) prompting~\citep{chainofthought} improves LLM reasoning by generating intermediate chains before generating the final output. This approach showed a great improvement for many reasoning tasks such as code generation where a complex problem benefits from its structured decomposition. Recent advances in structured CoT~\citep{structuredcot} demonstrate that organizing reasoning steps according to code structures significantly improves code generation quality. The key insight is that structured reasoning steps naturally align with the hierarchical nature of source code, leading to more coherent and syntactically correct outputs. Recent methods such as, Chain-of-Code~\citep{li2024chaincodereasoninglanguage} formats sub-tasks as flexible pseudo-code for iterative refinement, while least-to-most prompting~\citep{zhou2023leasttomostpromptingenablescomplex} tackles complex problems by decomposing them into sequential subproblems and SeqZero~\citep{yang2022seqzerofewshotcompositionalsemantic} addresses the compositional nature of formal languages by decomposing semantic parsing into sub-clause generation, avoiding the generation of lengthy canonical utterances in a single step. These approaches collectively demonstrate the effectiveness of structured reasoning in formal language generation.

Given the structured nature of constraint programming, we consider a CoT-based generation strategy well-suited for producing \textsc{MiniZinc} code. We hypothesize that aligning the generation process with the typical modeling workflow, progressing from parameters to variables, then constraints, and finally objectives facilitates more accurate outputs. To reinforce this alignment, we incorporate \textsc{MiniZinc}-specific modeling principles that help mitigate common errors such as missing bounds, inconsistent types, or misplaced declarations.
\subsection{Knowledge Graphs}
\label{sec:kg}
When interfacing with LLMs, providing structured information to these foundational, pre-trained models, typically used as part of in-context learning, have shown to be important~\citep{pankg2024}. A Knowledge Graphs (KGs) are particularly effective for representing information in a structured way. KGs are structured representations of information where entities are represented as \textit{nodes} and their relationships as \textit{edges}. KGs have become fundamental tools in various domains, including boosting Google's search engine~\citep{knowledgegraph} and organization of databases~\citep{dbpedia,yago,pubgraph}. In these graphs, real-world concepts and their relationships are captured in a format that both humans and machines can process effectively.

Knowledge graphs also serve as effective intermediate representations for improving language model reasoning and code generation. Chang et al. show that integrating commonsense knowledge graphs into pretrained language models significantly improves reasoning performance, with explicit attention mechanisms outperforming implicit integration methods~\citep{chang2021incorporatingcommonsenseknowledgegraph}. Similarly, dynamic knowledge graphs that capture story-specific facts have shown substantial improvements in narrative comprehension tasks~\citep{Andrus_Nasiri_Cui_Cullen_Fulda_2022}, enabling models to handle longer documents by creating information-rich prompts that supplement limited context windows. The knowledge prompting paradigm~\citep{wang2022knowledgepromptingpretrainedlanguage} further advances this approach by constructing domain-specific knowledge subgraphs and transforming them into natural language prompts, achieving superior performance across multiple natural language understanding tasks. This body of work demonstrates that structured knowledge representations can be effectively leveraged as intermediate steps in complex generation tasks.

Given the challenges large language models face when generating code for less commonly known languages, such as constraint modeling software we consider, structured knowledge graphs to be a particularly effective intermediate representation for \textsc{MiniZinc} code generation. The idea is to align the explicit nad relational nature of knowledge graphs with the declarative structure of \textsc{MiniZinc} models. Representing the problem at an atomic level through a graph helps preserve these dependencies and offers a stable representation for generation. To operationalize this, we construct knowledge graphs in the standard Turtle (TTL) format that explicitly capture the structure of the problem. For example, connecting the objective node to its relevant variables and reflect optimization dependencies. This structured representation, designed to imitate the logical structure of modeling code, facilitates a smoother transition from the unstructured text of a problem description to executable models. More concretely, in Appendix~\ref{appendix:knowledge_graph_generation}, we share an example knowledge graph of a combinatorial problem and the prompt used to obtain this knowledge graph representation.

\subsection{Grammar Constrained Generation}
\label{sec:grammar}
Lastly, grammar-constrained generation is an effective method for ensuring syntactic correctness in formal language generation. Willard et al. reformulate neural text generation as transitions between finite-state machine states~\citep{willard2023efficientguidedgenerationlarge}. This enables efficient guidance using regular expressions and context-free grammars through vocabulary indexing. This approach guarantees structural compliance while adding minimal overhead to the generation process. Recent advances in grammar-aligned decoding~\citep{park2024grammaraligneddecoding} address the distribution-distortion problem inherent in constrained decoding, where grammatical outputs may not reflect the LLM's true conditional probabilities. Their adaptive sampling approach ensures both grammatical correctness and alignment with the model's distribution. Similarly, XGrammar~\citep{dong2025xgrammarflexibleefficientstructured} achieves significant speedups by partitioning vocabulary into context-independent and context-dependent tokens, enabling efficient structured generation for applications that require parsed outputs. Frameworks like ReLM~\citep{kuchnik2023validatinglargelanguagemodels} demonstrate the validation capabilities of grammar-based approaches using regular expressions and achieving substantial efficiency gains in evaluation tasks. Practical implementations such as the Guidance framework\footnote{\url{https://github.com/guidance-ai/guidance}} provide programming paradigms for steering language models with structured output control through regex and context-free grammars. These developments collectively show that formal grammar specifications can effectively constrain language model outputs while maintaining generation quality.

\newpage

Given the practical constraints of using closed-source models such as GPT, which restrict access to token probabilities, and the performance limitations of open-source alternatives on \textsc{MiniZinc} generation, we consider grammar‑based validation a suitable post-processing strategy rather than relying on constrained decoding. The idea is to decouple syntax enforcement from generation enables us to leverage powerful LLMs while still ensuring syntactic correctness. By applying grammar‑driven validation after generation, we catch violations of \textsc{MiniZinc}’s formal syntax such as Boolean expression errors, missing delimiters without needing direct control over the decoding process. More concretely, in Appendix~\ref{appendix:grammar_example}, we share example grammar rules for \textsc{MiniZinc}.\footnote{We are grateful to Guido Tack from \textsc{MiniZinc} for providing us with the internal Bison parser output of the \textsc{MiniZinc} grammar.}

\section{Text2Model Translation Problem}
\label{sec:formulation}
We start by formalizing \textsc{Text2Model}, the problem of generating combinatorial models from textual descriptions.

Let $\mathbf{Input} = (description, parameters, output, metadata)$ represent the input specification where:
\begin{itemize}
    \item $description$: is the natural language description of the problem
    \item $parameters = \{p_1, ..., p_n\}$: is the set of input parameters where each $p_i = (d_i, s_i, \mathbf{h}_i)$ consists: 
        \begin{itemize}
            \item $d_i$: is the natural language definition of the parameter
            \item $s_i$: is the parameter symbol, e.g., `num\_warehouses', `num\_customers', `P'
            \item $h_i$: is the shape information of the parameter, e.g., scalar, vector, matrix, etc.
        \end{itemize}
    \item $output = \{o_1, ..., o_k\}$ is the set of output variables where each $o_i = (d_i, s_i, \mathbf{h}_i)$ consists: 
        \begin{itemize}
            \item $d_i$: is the natural language definition of the output
            \item $s_i$: output symbol (e.g., 'X', 'Y')
            \item $h_i$: is the shape information of the output, e.g., scalar, vector, matrix, etc.
        \end{itemize}
    \item $metadata$: metadata containing problem properties, e.g., application domain, objective type, constraint types
\end{itemize}

\noindent In addition, let $\mathbf{Data}$ be a concrete data instance specification, in our case a .dzn file for \textsc{MiniZinc}.

The goal of \textsc{Text2Model} is to learn a function $f: (Input, Data) \rightarrow \mathcal{M}$ where $\mathcal{M}$ represents the space of valid \textsc{MiniZinc} models. 

Given the input, $Input$, and instance data, $Data$, the function is aimed at generating a data-compatible model that correctly implements the given specification. Important to note that the complexity of the translation problem lies in what elements of the input is used by the translation function. The hardest case is to map directly from pure text, $description$, to a concrete model without depending on any $parameters$, $output$, and $metadata$ knowledge.

\begin{figure}[t]
    \centering
    \includegraphics[width=0.8\linewidth]{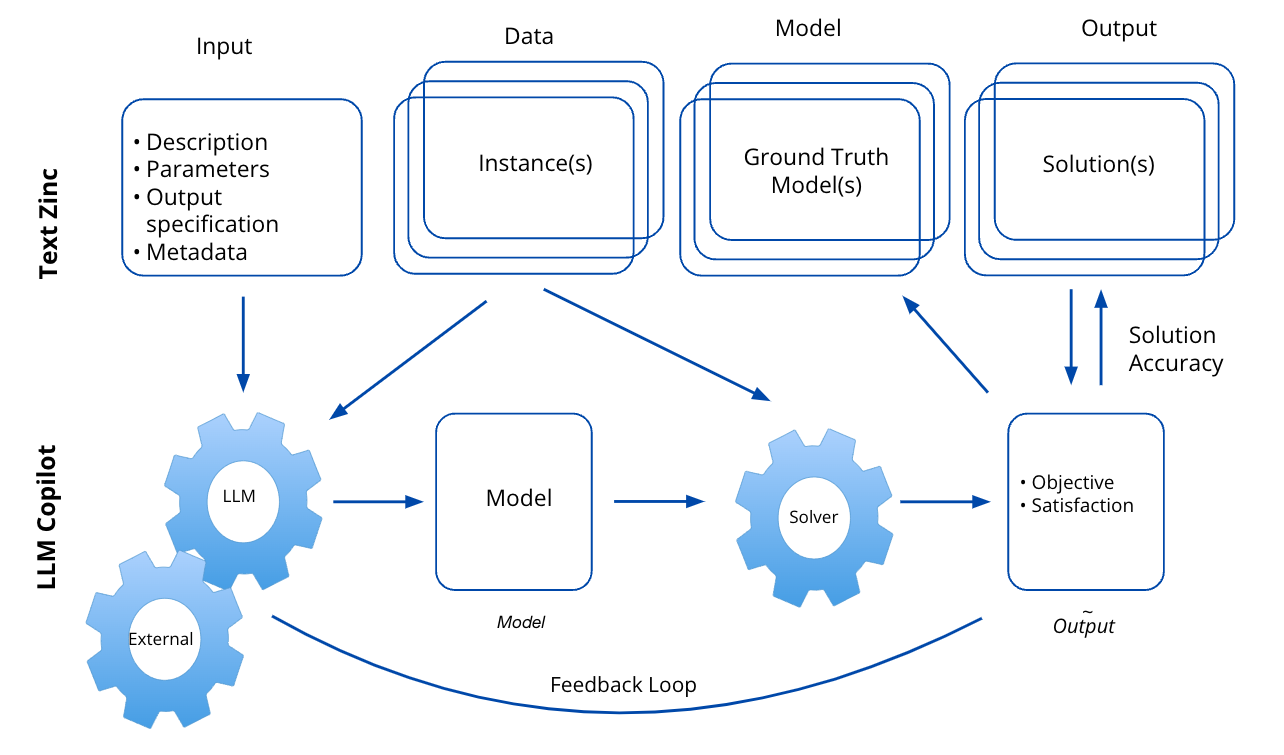}
    \caption{Text2Model copilots leveraging the Text2Zinc dataset.}
    \label{fig:architecture}
\end{figure}

\section{Text2Model copilots}
\label{sec:copilots}

Figure~\ref{fig:architecture} presents our  \textsc{Text2Model}\footnote{\url{https://github.com/skadio/text2model}} architecture for creating copilots to address the text-to-model translation problem. This blueprint is partially inspired by the Holy Grail 2.0 approach~\citep{tsouros2023holy}. We design various strategies that leverages the strengths of LLMs and intermediate representations. Our pipeline leverages the \textsc{Text2Zinc} dataset, which we explore in the next section.

At a high-level, given the input and data pair, and an LLM, potentially with external information (e.g., provided by a knowledge graph), our copilots generate a constraint model, which is then verified using a solver. The output is evaluated for execution and solution accuracy. The feedback loop enables multi-call strategies, where potential errors are resurfaced back to the LLM in subsequent cycles. 

We propose three categories of approaches: single-call strategies that generate complete \textsc{MiniZinc} models in one interaction (i.e. zero-shot), multi-call strategies that incorporate iterative refinement (i.e., few-shot), and compositional strategies that decompose the task into sequential sub-components (i.e., agentic) and combine result together.



\subsection{Single-Call Strategies}
These approaches generate complete \textsc{MiniZinc} models through a single LLM interaction.
\begin{enumerate}
   \item \textbf{Zero-Shot}: This baseline approach provides the problem description and expected input parameters and instructs the model to generate syntactically correct \textsc{MiniZinc} code.
   \item \textbf{Chain-of-Thought (CoT)}: This approach enhances the baseline strategy to enforce structured reasoning through a sequential chain-of-thought process, combining with general \textsc{MiniZinc} development guidelines covering variable bounds, constraint separation, proper declaration ordering and other general principles.
\end{enumerate}

\subsection{Multi-Call Strategies}
These approaches use multiple LLM interactions to iteratively refine the generated \textsc{MiniZinc} model.
\begin{enumerate}
   \item \textbf{Knowledge Graph}: As covered in~\ref{sec:kg}, this approach constructs a knowledge-graph style intermediate representation and leverages this representation to generate final \textsc{MiniZinc} code.
   \item \textbf{CoT + Code Validation}: Combines CoT approach with additional validation that ensures logical consistency between the generated model and problem requirements, verifying alignment of parameter names, constraints, and objective with the original problem description.
   \item \textbf{CoT + Grammar Validation}: As covered in~\ref{sec:grammar}, this approach combines CoT approach with syntactic error correction using formal \textsc{MiniZinc} grammar specifications to ensure strict adherence to syntax rules, covering model structure, type declarations, operators, and other \textsc{MiniZinc} grammar rules.
   \item \textbf{CoT + Code \& Grammar Validation}: Combines CoT with both code and grammar validation.
\end{enumerate}

\subsection{Agentic Strategies}
These approaches decompose model generation into sequential sub-tasks, generating parameters, variables, constraints, and objectives independently via specialized agents before combining them together. The idea of decomposition is based on the \textsc{Ner4Opt} framework~\citep{ner4opt2024}.

\begin{enumerate}
   \item \textbf{Agentic}: Parameters and decision variables, constraints, and objective function are generated independently, and the results are then combined to form the final MiniZinc model.
   \item \textbf{Agentic + Code Validation}: Combines the compositional approach with final code validation step.
\end{enumerate}

Overall, our design philosophy is to remain principled and explore strategies from simple single-call approaches to sophisticated multi-call and agentic workflows, each building upon previous insights to enhance model execution and solution accuracy. More concretely, in Appendix~\ref{appendix:appendix_prompts}, we document all our prompts for reference.


\section{Text2Zinc Dataset}
\label{sec:dataset}
We now introduce \textsc{Text2Zinc}\footnote{\url{https://huggingface.co/datasets/skadio/text2zinc}}, \textit{the first cross-domain and solver-agnostic dataset} that covers both satisfaction and optimization problems for text-to-model translation tasks. The dataset is designed to be solver-agnostic, thanks to integration with \textsc{MiniZinc} models. It currently comprises a total of 1,775 natural language problem instances, of which 110 have been carefully selected, manually verified, and augmented to serve as high-quality data. These verified instances are drawn from six primary sources that blends optimization and satisfaction problems from the following resources: \textsc{NLP4LP}~\citep{optimus}, \textsc{ComplexOR}~\citep{complexor}, \textsc{LPWP}~\citep{lpwp}, \textsc{CspLib}~\citep{csplib}, Hakank's collection~\citep{hakank}, \textsc{IndustryOR}, \textsc{MAMO (easy, complex)}, and \textsc{NL4Opt}~\citep{Huang_2025}.


\begin{figure*}[t]
    \centering
    \includegraphics[width=0.8\textwidth]{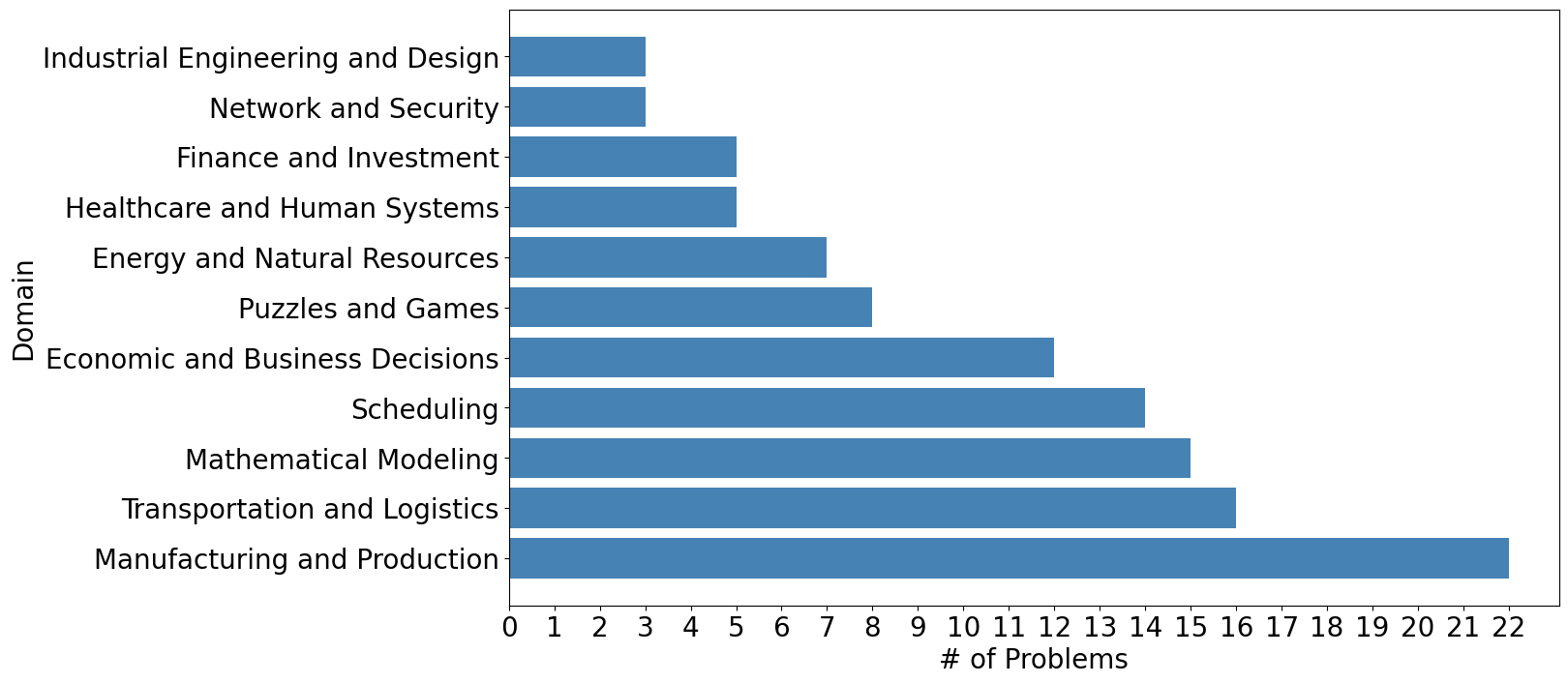}
    \caption{Distribution of problems across application domains.}
    \label{fig:domains}
\end{figure*}

From this larger collection, we carefully curated problems that clearly distinguish between data and parameter components. We standardize the diverse input formats (json, dzn, mzn, html, txt) to our unification format. To serve as supervised labels, we provide ground truth outputs for most problems to enable validation. Each problem instance is enriched with extensive metadata (e.g., application domain) to support diverse modeling approaches and future research needs (e.g., retrieval-augmented generation using in-context examples from the same domain). 

As shown in Figure~\ref{fig:domains}, the dataset spans 11 application domains and involves linear programming, integer programming, mixed-integer programming, and constraint programming problems. More details and dataset statistics are included in Appendix~\ref{appendix:text2zinc_stats}.

\subsection{Unification Schema}
\textsc{Text2Zinc} format unifies aforementioned dataset in a schema that divides each problem instance into four components: \emph{Input}, \emph{Data}, \emph{Model}, and \emph{Output}. The details of each component is as follows:

\begin{figure}[t]
    \centering
    \scalebox{0.6}
    {
\definecolor{promptbg}{RGB}{220,240,255}
\definecolor{outputbg}{RGB}{255,250,220}
\definecolor{bordercolor}{RGB}{100,100,100}
\definecolor{headerbg}{RGB}{150,180,220}
\definecolor{headerfg}{RGB}{255,255,255}

\lstset{
    basicstyle=\ttfamily\small,
    columns=fullflexible,
    breaklines=true
}

\tcbset{
    mybox2/.style={
        colframe=bordercolor,
        colback=promptbg,
        coltitle=headerfg,
        fonttitle=\bfseries,
        boxrule=0.5mm,
        width=25cm,
        rounded corners,
        enhanced,
        colbacktitle=headerbg,
        title style={left color=headerbg, right color=headerbg, rounded corners},
    },
    outputbox/.style={
        colframe=bordercolor,
        colback=outputbg,
        coltitle=headerfg,
        fonttitle=\bfseries,
        boxrule=0.5mm,
        width=23cm,
        rounded corners,
        enhanced,
        colbacktitle=headerbg,
        title style={left color=headerbg, right color=headerbg, rounded corners},
    }
}

\lstset{
    basicstyle=\ttfamily\small,
    columns=fullflexible,
    breaklines=true,
    tabsize=2, 
}

\begin{tcolorbox}[mybox2, title=input.json]

    \begin{lstlisting}
{
  "parameters": [
    {
      "definition": "Number of courses",
      "symbol": "courses",
      "shape": []
    },
    {
      "definition": "Number of periods",
      "symbol": "periods",
      "shape": []
    },
    {
      "definition": "Number of rooms available",
      "symbol": "rooms",
      "shape": []
    },
    {
      "definition": "Binary matrix where A[i,j]=1 indicates lectures of course i can be scheduled at period j",
      "symbol": "available",
      "shape": ["courses", "periods"]
    },
    {
      "definition": "Conflict matrix where M[i,j]=1 if courses i and j have common students",
      "symbol": "conflict",
      "shape": ["courses", "courses"]
    },
    {
      "definition": "Array containing the number of lectures required per course",
      "symbol": "requirement",
      "shape": ["courses"]
    }
  ],
  "output": [
    {
      "definition": "Timetable grid where 1 represents a scheduled lecture and 0 represents an unscheduled lecture",
      "symbol": "timetable",
      "shape": ["courses", "periods"]
    }
  ],
  "description": "Lecture timings need to be scheduled for courses across a limited number of periods. Each course requires a specific number of lectures and can only be assigned to certain periods due to availability constraints. Some courses have conflicts due to having common students and cannot be scheduled at the same time. Additionally, there is a limited number of rooms that can be used and thus a maximum number of lectures that can occur simultaneously. How can we allocate lectures to periods while ensuring all constraints are met?",
  "identifier": "or_lp_ip_scheduling_problem_2",
  "metadata": {
    "name": "Timetable Problem", "domain": "Scheduling", "objective": "satisfy", "source": "hakank", "constraints": [
      "forall", "<=", "+", "=", "sum"]
  }
}


\end{lstlisting}

\end{tcolorbox}}
    \caption{\footnotesize An example input with description, parameters, metadata, and output fields.}
    \label{fig:input}
\end{figure}



\noindent \textbf{Input:} As shown in Figure~\ref{fig:input}, a JSON file that encapsulates a comprehensive description of a problem instance. This file is organized into several key sections:

\begin{itemize}
    \item \textbf{Input - Problem Description:} A standardized natural language problem description of the problem. To maintain abstraction, we avoid including specific parameter names and values, instead incorporating them in the data file. This retains a neutral language in description. 
    \item \textbf{Input - Parameters:} Additional information about parameters  included in the \texttt{data.dzn} file. Each parameter is provided with a natural language explanation of its meaning and format, an associated symbol corresponding to its name in the \texttt{data.dzn} file, and a specification of its shape, which can be either an n-dimensional list or a scalar, represented by an empty list.
    \item \textbf{Input - Output Specification:} A detailed explanation of the expected output format, including the name and shape of any decision variables to be output. This is required to evaluate satisfaction problems.
    \item \textbf{Input - Metadata:} Additional contextual information, such as a descriptive problem title, a domain, and subdomain that enrich the problem, an objective (one of \texttt{satisfy}, \texttt{maximize}, or \texttt{minimize}), and a list of automatically generated keywords that reflect the constraints used in the problem, such as all\_different and $\leq$. This also includes a unique identifier which links the problem instance to its source.
\end{itemize}


\begin{figure}[t]
    \centering
    \scalebox{0.6}
    {
\definecolor{promptbg}{RGB}{220,240,255}
\definecolor{outputbg}{RGB}{255,250,220}
\definecolor{bordercolor}{RGB}{100,100,100}
\definecolor{headerbg}{RGB}{150,180,220}
\definecolor{headerfg}{RGB}{255,255,255}

\lstset{
    basicstyle=\ttfamily\small,
    columns=fullflexible,
    breaklines=true
}

\tcbset{
    mybox2/.style={
        colframe=bordercolor,
        colback=promptbg,
        coltitle=headerfg,
        fonttitle=\bfseries,
        boxrule=0.5mm,
        width=24cm,
        rounded corners,
        enhanced,
        colbacktitle=headerbg,
        title style={left color=headerbg, right color=headerbg, rounded corners},
    }
}

\lstset{
    basicstyle=\ttfamily\small,
    columns=fullflexible,
    breaklines=true,
    tabsize=2, 
}

\begin{tcolorbox}[mybox2, title=model.mzn]

    \begin{lstlisting}
include "globals.mzn";

% Input parameters
int: courses;
int: periods;
int: rooms;

array[1..courses, 1..periods] of int: available;
array[1..courses, 1..courses] of int: conflict;
array[1..courses] of int: requirement;

% Decision variables
array[1..courses, 1..periods] of var 0..1: timetable;

% Solve
solve :: int_search(
    [timetable[c, p] | c in 1..courses, p in 1..periods],
    most_constrained,
    indomain_split,
    complete
) satisfy;

% Constraints
constraint
    % 1. Conflicts: Courses with common students must not be scheduled at the same time
    forall(c1, c2 in 1..courses where c1 < c2) (
        if conflict[c1, c2] = 1 then
            forall(p in 1..periods) (
                timetable[c1, p] + timetable[c2, p] <= 1
            )
        else
            true
        endif
    )
    % 2. Availabilities: Courses can only be scheduled in available periods
    /\
    forall(c in 1..courses, p in 1..periods) (
        if available[c, p] = 0 then
            timetable[c, p] = 0
        else
            true
        endif
    )
    % 3. Rooms: At most `rooms` lectures can be scheduled per period
    /\
    forall(p in 1..periods) (
        sum([timetable[c, p] | c in 1..courses]) <= rooms
    )
    % 4. Number of lectures per course must match the requirement
    /\
    forall(c in 1..courses) (
        sum([timetable[c, p] | p in 1..periods]) = requirement[c]
    );
\end{lstlisting}

\end{tcolorbox}}
    \caption{\footnotesize An example \textsc{MiniZinc} model.}
    \label{fig:minizinc}
\end{figure}

\medskip
\noindent \textbf{Model:} As shown in Figure~\ref{fig:minizinc}, an MZN file containing the \textsc{MiniZinc} model file that formulates the problem and serves as a verifier for satisfaction problems. The model can be used as a verifier by treating the output from the LLM-generated code in DZN format as input to the model. The constraints are decomposed where possible for better clarity. 





\newpage
\noindent \textbf{Data:} As shown in Figure~\ref{fig:data}, a DZN file representing a concrete problem instance. This instance data is used to validate the code generated by the language model in conjunction with the ground truth output labels. 


\medskip
\noindent \textbf{Output:} As shown in Figure~\ref{fig:output}, a JSON file containing the assignments of variables specified in the objective values for a problem. Two essential elements of the output are the values of variables (serving as a feasible solution for satisfaction problems) and the optimal value (for optimization problems). 

\begin{figure}[t]
  \centering
  \begin{subfigure}{0.48\linewidth}
    \centering
    \resizebox{\linewidth}{!}{
\definecolor{promptbg}{RGB}{220,240,255}
\definecolor{outputbg}{RGB}{255,250,220}
\definecolor{bordercolor}{RGB}{100,100,100}
\definecolor{headerbg}{RGB}{150,180,220}
\definecolor{headerfg}{RGB}{255,255,255}

\lstset{
    basicstyle=\ttfamily\small,
    columns=fullflexible,
    breaklines=true
}

\tcbset{
    mybox/.style={
        colframe=bordercolor,
        colback=promptbg,
        coltitle=headerfg,
        fonttitle=\bfseries,
        boxrule=0.5mm,
        width=9cm,
        rounded corners,
        enhanced,
        colbacktitle=headerbg,
        title style={left color=headerbg, right color=headerbg, rounded corners},
    },
    outputbox/.style={
        colframe=bordercolor,
        colback=outputbg,
        coltitle=headerfg,
        fonttitle=\bfseries,
        boxrule=0.5mm,
        width=8cm,
        rounded corners,
        enhanced,
        colbacktitle=headerbg,
        title style={left color=headerbg, right color=headerbg, rounded corners},
    }
}

\tcbset{
    mybox2/.style={
        colframe=bordercolor,
        colback=promptbg,
        coltitle=headerfg,
        fonttitle=\bfseries,
        boxrule=0.5mm,
        width=16cm,
        rounded corners,
        enhanced,
        colbacktitle=headerbg,
        title style={left color=headerbg, right color=headerbg, rounded corners},
    },
    outputbox/.style={
        colframe=bordercolor,
        colback=outputbg,
        coltitle=headerfg,
        fonttitle=\bfseries,
        boxrule=0.5mm,
        width=16cm,
        rounded corners,
        enhanced,
        colbacktitle=headerbg,
        title style={left color=headerbg, right color=headerbg, rounded corners},
    }
}

\lstset{
    basicstyle=\ttfamily\small,
    columns=fullflexible,
    breaklines=true,
    tabsize=2, 
}

\begin{tcolorbox}[mybox, title=data.dzn]

    \begin{lstlisting}
int: courses = 5;
int: periods = 20;
int: rooms = 2;
array[1..courses, 1..periods] of int: available = array2d(1..courses, 1..periods, [
    %  1  2  3  4  5  6  7  8  9  0  1  2  3  4  5  6  7  8  9  0
       0, 0, 1, 1, 1, 1, 1, 1, 1, 1, 1, 1, 0, 1, 1, 0, 1, 1, 1, 1,
       1, 1, 0, 0, 1, 0, 1, 1, 0, 1, 1, 1, 1, 1, 1, 1, 1, 1, 1, 1,
       0, 0, 0, 1, 1, 1, 1, 0, 1, 1, 1, 1, 0, 1, 1, 1, 1, 0, 1, 1,
       1, 1, 1, 0, 0, 0, 1, 1, 1, 1, 1, 1, 1, 1, 1, 1, 1, 0, 1, 1,
       1, 1, 1, 1, 1, 1, 1, 1, 1, 1, 1, 1, 1, 1, 1, 1, 1, 1, 1, 1
]);
array[1..courses, 1..courses] of int: conflict = array2d(1..courses, 1..courses, [
    % Conflict matrix
    0, 1, 0, 0, 1,
    1, 0, 0, 1, 0,
    0, 0, 0, 0, 1,
    0, 1, 0, 0, 1,
    1, 0, 1, 1, 0
]);
array[1..courses] of int: requirement = [6, 10, 14, 6, 4];


\end{lstlisting}

\end{tcolorbox}}
    \captionsetup{font=footnotesize}
    \caption{An example data instance.}
    \label{fig:data}
  \end{subfigure}
  \hfill                    
  \begin{subfigure}{0.48\linewidth}
    \centering
    \resizebox{\linewidth}{!}{
\definecolor{promptbg}{RGB}{220,240,255}
\definecolor{outputbg}{RGB}{255,250,220}
\definecolor{bordercolor}{RGB}{100,100,100}
\definecolor{headerbg}{RGB}{150,180,220}
\definecolor{headerfg}{RGB}{255,255,255}

\lstset{
    basicstyle=\ttfamily\small,
    columns=fullflexible,
    breaklines=true
}

\tcbset{
    mybox/.style={
        colframe=bordercolor,
        colback=promptbg,
        coltitle=headerfg,
        fonttitle=\bfseries,
        boxrule=0.5mm,
        width=9cm,
        rounded corners,
        enhanced,
        colbacktitle=headerbg,
        title style={left color=headerbg, right color=headerbg, rounded corners},
    },
    outputbox/.style={
        colframe=bordercolor,
        colback=outputbg,
        coltitle=headerfg,
        fonttitle=\bfseries,
        boxrule=0.5mm,
        width=8cm,
        rounded corners,
        enhanced,
        colbacktitle=headerbg,
        title style={left color=headerbg, right color=headerbg, rounded corners},
    }
}

\tcbset{
    mybox2/.style={
        colframe=bordercolor,
        colback=promptbg,
        coltitle=headerfg,
        fonttitle=\bfseries,
        boxrule=0.5mm,
        width=16cm,
        rounded corners,
        enhanced,
        colbacktitle=headerbg,
        title style={left color=headerbg, right color=headerbg, rounded corners},
    },
    outputbox/.style={
        colframe=bordercolor,
        colback=outputbg,
        coltitle=headerfg,
        fonttitle=\bfseries,
        boxrule=0.5mm,
        width=16cm,
        rounded corners,
        enhanced,
        colbacktitle=headerbg,
        title style={left color=headerbg, right color=headerbg, rounded corners},
    }
}

\lstset{
    basicstyle=\ttfamily\small,
    columns=fullflexible,
    breaklines=true,
    tabsize=2, 
}

\begin{tcolorbox}[mybox, title=output.json]
    \begin{lstlisting}
{
"timetable": [
    [0, 0, 1, 1, 0, 1, 0, 1, 0, 0, 0, 1, 0, 1, 0, 0, 0, 0, 0, 0],
    [1, 1, 0, 0, 1, 0, 0, 0, 0, 0, 0, 0, 1, 0, 1, 1, 1, 1, 1, 1],
    [0, 0, 0, 1, 1, 1, 1, 0, 1, 1, 1, 1, 0, 1, 1, 1, 1, 0, 1, 1],
    [0, 0, 1, 0, 0, 0, 1, 1, 1, 1, 1, 0, 0, 0, 0, 0, 0, 0, 0, 0],
    [1, 1, 0, 0, 0, 0, 0, 0, 0, 0, 0, 0, 1, 0, 0, 0, 0, 1, 0, 0]
  ]
  }

    \end{lstlisting}
\end{tcolorbox}}
    \captionsetup{font=footnotesize}
    \caption{Model execution output for the instance.}
    \label{fig:output}
  \end{subfigure}

  \caption{A data instance (left) and the corresponding output produced by the \textsc{MiniZinc} model (right).}
  \label{fig:data-output}
\end{figure}
\medskip

\subsection{Dataset Verification}
Input parameters are validated against the DZN files through automated verification. The output files for instances with model files contain the complete \textsc{MiniZinc} model results in JSON format, while instances without models include only the objective value. Model files are provided for all satisfaction problems, and all included models have been verified for successful compilation.


\begin{figure}[t]
    \centering
    \includegraphics[width=\linewidth]{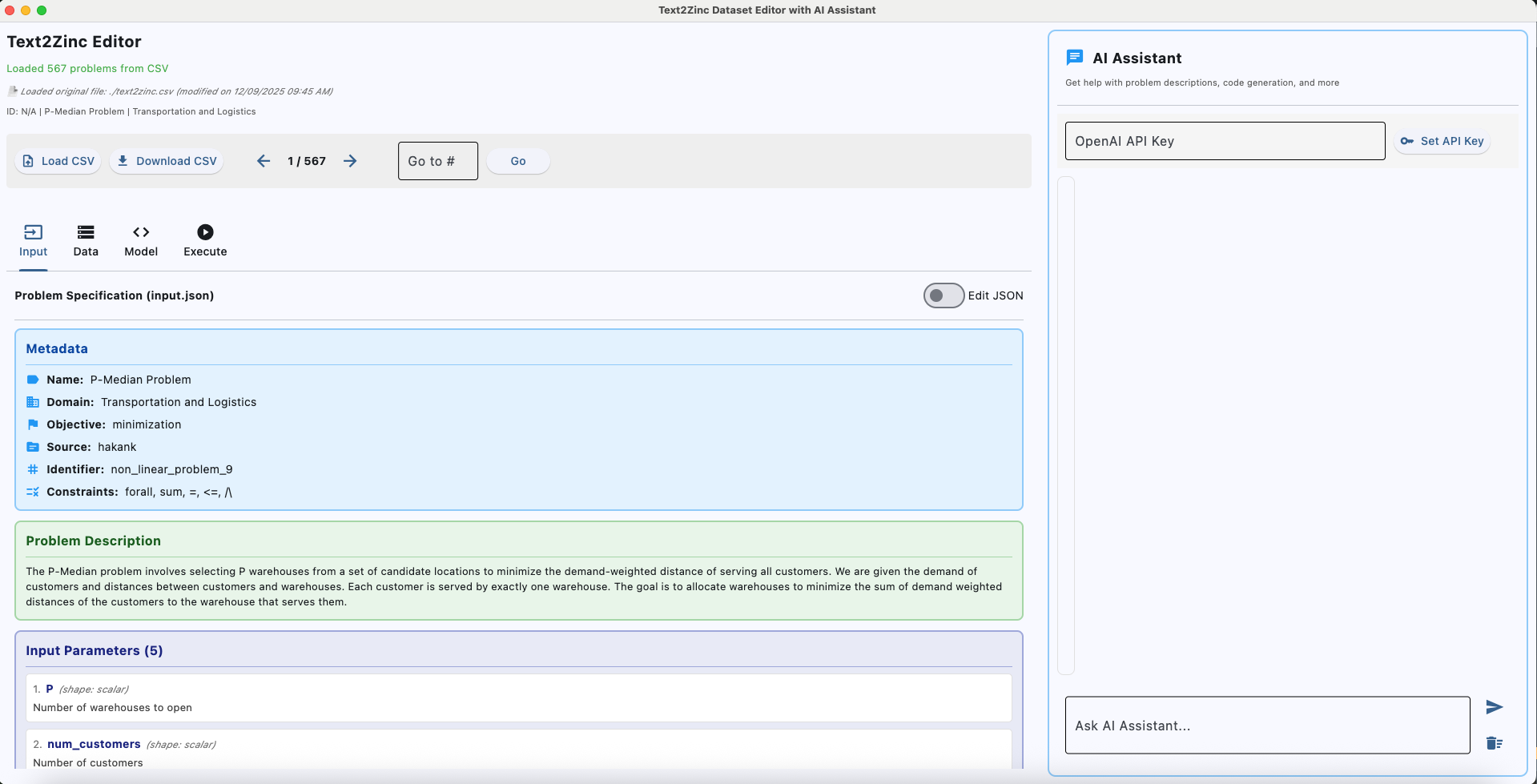}
    \vspace{0.2cm}
    \caption{\textsc{Text2Zinc} Editor for dataset curation and validation with built-in AI Assistant.}
    \label{fig:editor}
    \vspace{0.2cm}
\end{figure}

\newpage
\subsection{Interactive Editor}
To support ongoing data curation, we accompany \textsc{Text2Zinc} with the open-source \textsc{Text2Zinc Editor}\footnote{\url{https://huggingface.co/spaces/skadio/text2zinc-editor}}. As shown in Figure~\ref{fig:editor}, the editor provides an interface for browsing, validating, and refining problems in the dataset (displayed on the left) with a built-in AI assistant (displayed on the right).

The left panel provides access to different problem components. The \textit{Input} tab displays the problem description, metadata (name, domain, objective type), input parameters, and expected output variables in a structured view. Users can also switch to a raw JSON editing mode if needed. 

The \textit{Data} tab shows the \texttt{data.dzn} file containing the instance data, while the \textit{Model} tab contains the ground truth \textsc{MiniZinc} model. The \textit{Execute} tab allows users to run the model against the data using different solvers (Gecode, Chuffed, HiGHS, OR-Tools, etc.) with configurable timeouts, and compare the execution results against the expected output. This makes it easy to verify that the model produces correct results. A verification checkbox lets annotators mark problems as validated, and all edits can be saved back to the dataset.

The right panel provides an optional chat assistant powered by OpenAI. Users can provide their API key to interact with the assistant, which automatically has access to the current problem context (description, data, and model code). This helps refining descriptions, alternative instance generation, and drafting initial model code. The assistant maintains the conversation history to enable iterative refine through dialogue.

\section{Computational Experiments}
\label{sec:experiments}
So far we covered the \textsc{Text2Model} problem description, different copilot approaches, and the \textsc{Text2Zinc} dataset. Next, we conduct computational experiments to test the performance of LLMs for generating \textsc{MiniZinc} models on this dataset. The goal of our experiments is to establish \textit{baseline performance} to serve as a lower bound on this task to inspire new methodologies. In addition, we compare our copilot strategies with the latest work in the literature, including \textsc{Gala} Global Agents~\cite{cai2025galagloballlmagents}, \textsc{Orlm}~\cite{orlm}, and \textsc{OptiMind}~\cite{optimind} from Microsoft. 

\subsection{Evaluation Metrics}
Given a dataset of $N$ problem instances, each with a ground truth model $m^*_i$ and an LLM-generated model $m_i$, we define two key metrics:

\begin{enumerate}
    \item \textbf{Execution Accuracy} ($E_{acc}$): Measures the proportion of generated models that successfully compile and execute, where $\mathbb{I}(\cdot)$ is the indicator function.
    \begin{equation}
        E_{acc} = \frac{1}{N}\sum_{i=1}^N \mathbb{I}(execute(m_i) = \text{success})
        \label{eq:execution_accuracy}
    \end{equation}
    
    \item \textbf{Solution Accuracy} ($S_{acc}$): Measures the proportion of models that achieve the same objective value as the ground truth,
    where $obj(\cdot)$ returns the objective value and $m^*_i$ is the ground truth model for instance $i$.
    \begin{equation}
    S_{acc} = \frac{1}{N}\sum_{i=1}^N \mathbb{I}(obj(m_i) = obj(m^*_i))
    \label{eq:solution_accuracy}
    \end{equation}    
\end{enumerate}

\begin{table}[t]
\centering
\caption{\textsc{Text2Model} Benchmark with GPT-5.2 (Part - I)}
\label{tab:experimental-results-gpt52}
\begin{adjustbox}{max width=\textwidth}
\begin{tabular}{@{}lc|cc|cc|cc|cc|cc@{}}
\toprule
 &  & \multicolumn{2}{c|}{\textsc{NLP4LP} (n=65)} 
 & \multicolumn{2}{c|}{\textsc{ComplexOR} (n=7)} 
 & \multicolumn{2}{c|}{\textsc{LPWP} (n=5)} 
 & \multicolumn{2}{c|}{\textsc{CspLib} (n=11)} 
 & \multicolumn{2}{c}{\textsc{Hakank} (n=22)} \\
copilot Strategy & LLM Calls 
 & $E_{acc}$ & $S_{acc}$ 
 & $E_{acc}$ & $S_{acc}$ 
 & $E_{acc}$ & $S_{acc}$ 
 & $E_{acc}$ & $S_{acc}$ 
 & $E_{acc}$ & $S_{acc}$ \\
\midrule
\textsc{Zero Shot} & 1 
 & 63.08 & 32.31 
 & \best{100.0} & \best{57.14} 
 & \best{100.0} & \best{100.0} 
 & 63.64 & 36.36 
 & 50.00 & 31.82 \\
\textsc{Chain-of-Thought (CoT)} & 1 
 & 70.77 & 35.38 
 & \best{100.0} & 42.86 
 & \best{100.0} & \best{100.0} 
 & 54.55 & 27.27 
 & 59.09 & 36.36 \\
\textsc{Knowlege Graph} & 2 
 & 52.31 & 32.31 
 & 71.43 & \best{57.14} 
 & \best{100.0} & \best{100.0} 
 & 36.36 & 27.27 
 & 68.18 & 31.82 \\
\textsc{CoT + Code Validation} & 2 
 & 64.62 & 30.77 
 & 71.43 & 42.86 
 & \best{100.0} & \best{100.0} 
 & 72.73 & 36.36 
 & 63.64 & 36.36 \\
\textsc{CoT + Grammar} & 2 
 & \best{76.92} & \best{38.46} 
 & 85.71 & \best{57.14} 
 & \best{100.0} & \best{100.0} 
 & \best{90.91} & \best{45.45} 
 & \best{72.73} & 36.36 \\
\textsc{CoT + Code \& Grammar} & 3 
 & 69.23 & 32.31 
 & 85.71 & \best{57.14} 
 & \best{100.0} & 80.00 
 & \best{90.91} & \best{45.45} 
 & \best{72.73} & \best{45.45} \\
\textsc{Agentic} & 4 
 & 50.77 & 32.31 
 & 85.71 & \best{57.14} 
 & 60.00 & 60.00 
 & 63.64 & 18.18 
 & 54.55 & 31.82 \\
\textsc{Agentic + Code} & 5 
 & \best{76.92} & 29.23 
 & 85.71 & \best{57.14} 
 & \best{100.0} & \best{100.0} 
 & 81.82 & 36.36 
 & 68.18 & 36.36 \\
\textsc{Gala (\cite{cai2025galagloballlmagents})} & 5 
 & 66.15 & 32.31 
 & 85.71 & \best{57.14} 
 & \best{100.0} & \best{100.0} 
 & 63.64 & \best{45.45} 
 & 59.09 & 36.36 \\
\bottomrule
\end{tabular}
\end{adjustbox}
\end{table}

\subsection{Comparison of LLM copilots}
\label{sec:results}
Table~\ref{tab:experimental-results-gpt52} compares the solution and execution accuracy of various copilot strategies (\S\ref{sec:copilots}) on \textsc{Text2Zinc} dataset (\S\ref{sec:dataset}) in the order of LLM calls required using GPT-5.2. The immediate observation is, apart from the \textsc{LPWP} instances, LLM (GPT-5.2) strategies face extreme difficulty in text-to-model translation. The best solution accuracy across all approaches ranges between 38.46\% to 57.14\%, which clearly indicates that \textsc{Text2Zinc} remains an open benchmark for the community. Overall, our novel grammar encoding performs the best hinting at the potential of structured output generation.

\begin{table}[b]
\centering
\caption{\textsc{Text2Model} Benchmark with GPT-5.2 (Part - II)}
\label{tab:experimental-results-orlm-gpt52}
\begin{adjustbox}{max width=\textwidth}
\begin{tabular}{@{}lc|cc|cc|cc|cc@{}}
\toprule
 &  & \multicolumn{2}{c|}{\textsc{IndustryOR} (n=100)}
 & \multicolumn{2}{c|}{\textsc{Mamo-Easy} (n=652)}
 & \multicolumn{2}{c|}{\textsc{Mamo-Complex} (n=211)}
 & \multicolumn{2}{c}{\textsc{NL4Opt} (n=245)} \\
copilot Strategy & LLM Calls
 & $E_{acc}$ & $S_{acc}$
 & $E_{acc}$ & $S_{acc}$
 & $E_{acc}$ & $S_{acc}$
 & $E_{acc}$ & $S_{acc}$ \\
\midrule
\textsc{Zero Shot} & 1
 & 66.00 & 46.00
 & 95.71 & 79.75
 & 48.82 & 27.96
 & 93.88 & 74.29 \\
\textsc{Chain-of-Thought (CoT)} & 1
 & 80.00 & 49.00
 & 98.62 & 83.90
 & 57.35 & 39.34
 & \best{98.78} & \best{78.37} \\
\textsc{CoT + Code Validation} & 2
 & 86.00 & 49.00
 & 99.54 & 82.98
 & 94.31 & 54.98
 & 97.14 & 73.88 \\
\textsc{CoT + Grammar} & 2
 & \best{92.00} & 56.00
 & \best{99.85} & 83.74
 & 95.26 & 54.03
 & 97.96 & 77.96 \\
\textsc{CoT + Code \& Grammar} & 3
 & \best{92.00} & 55.00
 & \best{99.85} & 83.13
 & \best{96.68} & \best{55.92}
 & \best{98.78} & 77.55 \\
\textsc{Agentic} & 4
 & 74.00 & 48.00
 & 98.01 & 83.90
 & 62.56 & 38.86
 & 95.92 & 77.55 \\
\textsc{Agentic + Code} & 5
 & 86.00 & \best{57.00}
 & 99.23 & \best{84.97}
 & 92.89 & 54.03
 & 97.55 & 77.96 \\
\textsc{Gala (\cite{cai2025galagloballlmagents})} & 5
 & 77.00 & 52.00
 & 96.63 & 80.21
 & 72.04 & 41.71
 & 90.61 & 71.02 \\
\bottomrule
\end{tabular}
\end{adjustbox}
\end{table}

\begin{table}[t]
\centering
\caption{\textsc{Text2Model Benchmark with GPT-5.2 (Total n=1318)}}
\label{tab:total}
\begin{adjustbox}{max width=\textwidth}
\begin{tabular}{@{}l|cccccccc@{}}
\toprule
 & \textsc{Zero Shot} 
 & \textsc{CoT} 
 & \textsc{CoT + Code}
 & \textsc{CoT + Grammar}
 & \textsc{CoT + Code \& Grammar}
 & \textsc{Agentic}
 & \textsc{Agentic + Code}
 & \textsc{Gala} \\
\midrule
$E_{acc}$ 
 & 1094 & 1163 & 1246 & \best{1271} & \best{1271} & 1141 & 1253 & 1155 \\
$S_{acc}$ 
 & 848 & 913 & 927 & \best{956} & 949 & 904 & \best{956} & 880 \\
\bottomrule
\end{tabular}
\end{adjustbox}
\end{table}

Table~\ref{tab:experimental-results-orlm-gpt52} presents further results on the remaining \textsc{Text2Zinc} datasets. Again, the best solution accuracy across all approaches ranges from 57\% to 84.97\%, leaving significant room for improvement. Our grammar strategy achieves several best results and stands out as the most competitive approach overall. 

Table~\ref{tab:total} combines Table~\ref{tab:experimental-results-gpt52} and~\ref{tab:experimental-results-orlm-gpt52} to present the number of instances (out of 1318 in total) accurately modeled and solved by each copilot. Overall, \textsc{CoT} with our novel \textsc{Grammar} encoding and perform the best while 362 (27\%) instances remain out of LLM's reach.

\newpage
\subsection{Comparison with the State-of-the-Art: \textsc{Gala}, \textsc{Orlm}, and \textsc{OptiMind}}
We now compare our copilot strategies with the latest work in the literature. We consider \textsc{Gala} Global Agents~\cite{cai2025galagloballlmagents}, \textsc{Orlm}~\cite{orlm}, and \textsc{OptiMind}~\cite{optimind} from Microsoft. 
\textsc{Gala} is a recent agentic framework that focuses on decomposing the problem structure to associate specialized agents for global constraints. \textsc{Orlm} and \textsc{OptiMind} \textsc{Gala} are specialized LLMs that are fine-tuned specifically for text-to-model translation. Previously, \textsc{Orlm} has been shown to perform better than tag-BART~\cite{kani2022tagged}, Reflexion~\cite{shinn2023reflexion}, Chain-of-Experts~\cite{complexor}, and OptiMUS~\cite{optimus}. 

We re-run \textsc{Gala} and \textsc{OptiMind}, and for \textsc{Orlm}, we use the results from~\cite{orlm} since they also report performance on the same instances with the same solution accuracy metric.

\begin{table}[b]
\centering
\caption{Comparison with \textsc{Orlm} results from the original paper (\cite{orlm})}
\label{tab:orlm}
\begin{tabular}{@{}lcccc@{}}
\toprule
& \textsc{IndustryOr} & \textsc{Mamo-Easy} & \textsc{Mamo-Complex} & \textsc{Nl4Opt} \\
\midrule
\textsc{Orlm} Pass@1 & 38.0\% & 82.3\% & 37.4\% & \best{85.7\%} \\
Our \textsc{CoT} Pass@1 (from Table~\ref{tab:experimental-results-orlm-gpt52}) & \best{49.0\%} & \best{83.90\%} & \best{39.34\%} & 78.37\% \\
\midrule
\textsc{Orlm} Pass@2 & 40.0\% & 83.7\% & 49.8\% & 88.6\% \\
\textsc{Orlm} Pass@4 & 44.0\% & 85.9\% & 56.9\% & 91.4\% \\
\textsc{Orlm} Pass@8 & 49.0\% & \best{88.4\%} & \best{72.1\%} & \best{93.0\%} \\
\midrule

Our Best Pass@5 (from Table~\ref{tab:experimental-results-orlm-gpt52}) & \best{57.0\%} & 84.97\% & 55.92\% & 78.37\% \\
\bottomrule
\end{tabular}
\end{table}

\noindent \textbf{Comparison with \textsc{Gala}}: As shown in Table~\ref{tab:experimental-results-gpt52} and ~\ref{tab:experimental-results-orlm-gpt52}, \textsc{Gala} (\cite{cai2025galagloballlmagents}) performs reasonable well but still behind our best results. It performs better than our \textsc{Agentic} framework in \textsc{CspLib} and \textsc{Hakank} instances, where global constraints matter for constraint programming. This indicates the value of agentic decompositions that match the underlying modeling paradigm. 

\noindent \textbf{Comparison with \textsc{Orlm}}: Table~\ref{tab:orlm} compares \textsc{Orlm} (\cite{orlm}) with our copilots. This comparison is intended to be \textit{directional only} due the different nature of copilots and LLMs. \textsc{Orlm} is an instruct-tuned LLM for text-to-model translation and is stochastic in nature. \cite{orlm} reports the best solution accuracy out of several runs (Pass@K). Keep in mind that \textsc{Orlm} is specific to a single paradigm (mathematical programming) and a single solver (Cardinal Optimizer) whereas our approach is paradigm- and solver-agnostic. \textsc{Orlm} reports solution accuracy on \textsc{IndustryOr}, \textsc{Mamo-Easy}, \textsc{Mamo-Complex}, \textsc{Nl4Opt} datasets by generating Cardinal Optimizer models. Our copilots generate \textsc{MiniZinc} models for the same instances, which are now unified under \textsc{Text2Zinc}.

For Pass@1, our out-of-the-box \textsc{CoT} approach performs better than \textsc{Orlm} on these datasets except \textsc{Nl4Opt}. At a maximum, our copilots use 5 LLM calls, which can be directionally compared to \textsc{Orlm} Pass@4 and Pass@8. \textsc{Orlm} Pass@8 achieves the best results on \textsc{Mamo} and \textsc{Mamo-Complex} while our \textsc{Agentic + Code} @5 achieves the best in \textsc{IndustryOr}. Independent of approaches, \textsc{IndustryOr} and \textsc{Mamo-Complex} remain open for considerable improvement.

\noindent\textbf{Comparison with \textsc{OptiMind}}: We evaluated our co‑pilots in an ``as‑is'' manner by replacing GPT‑5.2 with \textsc{OptiMind}'s fine-tuned GPT-OSS-20b. \textsc{OptiMind} exhibited consistently poor performance (low single‑digit accuracy), so we omit detailed results here and report them in Appendix~\ref{appendix:optimind}. We note that \textsc{OptiMind} fine‑tunes GPT-OSS-20b specifically to generate \textsc{GurobiPy} models, and may therefore require re‑designing our co‑pilots and prompts to better align with \textsc{Gurobi}. Our goal in this paper is to remain paradigm‑ and solver‑agnostic, hence, we do not pursue such adaptations here. Interestingly, \textsc{Gala} (\cite{cai2025galagloballlmagents}), which also employs GPT‑OSS‑20b, reports competitive performance when generating \textsc{MiniZinc} models. Our findings suggest that \textsc{OptiMind}'s fine‑tuning GPT‑OSS‑20b to specialize on \textsc{Gurobi} may degrade its broader modeling capabilities. This highlights a potential caveat of task‑specific fine‑tuning where gains in one specialization may come at the cost of diminished general‑purpose performance.

We conduct additional experiments using both a lower‑cost model (GPT‑4) and a reasoning‑optimized model (GPT‑4o), with results presented in Appendix~\ref{appendix:gpt4_results}. For transparency into model limitations, we also include an error analysis and illustrative failure cases in Appendix~\ref{appendix:error_analysis}. 

Finally, we establish an online leaderboard for text-to-model translation and results presented in this paper form the basis of our HuggingFace \textsc{Text2Model} Leaderboard\footnote{\url{https://huggingface.co/spaces/skadio/text2model-leaderboard}}. This leaderboard is open to submissions and we welcome new approaches from the community to close the  performance gap observed in current LLMs.

\section{Related Work}
\label{sec:related}
There is growing interest in leveraging LLMs for optimization tasks to transform the way that decision makers interact with solvers in the form of a copilot system~\citep{SimchiLevi2025Democratizing,Wasserkrug2025,tsouros2023holy}. In the following, we review related work in this area.

\paragraph{From Natural Language to Optimization Models:}
Earlier systems relied on rule-based parsing to construct mixed-integer or logic models (e.g., \textsc{LGPSolver} for logic-grid puzzles~\citep{jabrayilzade2020lgpsolver} and \textsc{AutoLP} for linear programs~\citep{islam2021autoLP}).  The \textsc{Nl4Opt} competition~\citep{lpwp} formalized the task as a two-step pipeline: named-entity recognition followed by code generation.  Follow-up works such as \textsc{LaTeX2Solver} \citep{ramamonjison2023latex2solver} extended parsing to mathematical documents, while the ``Holy Grail 2.0'' blueprint~\citep{tsouros2023holy} envisioned conversational assistants that refine models interactively. 

\paragraph{Prompt-Driven and Learning-based Copilots:}
To get better results, researchers began use more sophisticated strategies such as structured prompts and retrieval, along with learning and fine-tuning to generate models from free form natural language text. The \textsc{Ner4Opt} line of works \citep{ner4opt2023,ner4opt2024} showed that fine-tuning transformers architecture on optimization specific corpora and inline entity tags boost the accuracy of generating correct \textsc{MiniZinc} models. Retrieval-augmented prompting specific to Constraint Programming (CP) settings has also proven effective~\citep{michailidis_et_al:LIPIcs.CP.2024.20} as well as in Prolog~\cite{DI2025114140}.

Modular agent pipelines push this further: \textsc{ComplexOR} employs a chain-of-experts architecture for difficult OR problems~\citep{complexor}, while \textsc{OptiMUS} decomposes formulation, debugging, and solving into separate GPT agents~\citep{optimus}. \textsc{Lean-Llm-Opt} focuses on the problem instances with relatively larger number of variables~\citep{liang2026largescaleoptimizationmodelautoformulation}. The recent \textsc{Gala} framework builds global agents for CP~\citep{cai2025galagloballlmagents}. Fine-tuning approaches include \textsc{Ner4Opt}~\citep{ner4opt2024}, \textsc{Orlm}~\citep{orlm}, \textsc{LLMOPT}~\citep{jiang2025llmopt}. Reinforcement Learning has also been used for generating structured queries from natural language~\citep{zhong2017seq2sql}. Although promising, these methods either focus on either one of optimization or satisfaction problems, and are often relegated to a single solver backend.

\paragraph{Integrating LLMs During the Solve Loop:}
Beyond model generation, there has been cross-pollination at a lower level between LLMs and solvers. Examples include \textsc{MiniZinc} streamliner generation \citep{streamllm}, CP–LLM co-processing via the MCP protocol \citep{szeider2025mcpsolverintegratinglanguagemodels}, hybrid SAT local-search guidance \citep{schidler2025extractingproblemstructurellms}, an API language allowing LLMs to communicate with constraint solvers~\citep{kesseli2025logicpy}, learning natural language interfaces with language models~\citep{dong2019thesis}, pre-training langauge models on scientific text ~\citep{wang2022knowledgepromptingpretrainedlanguage}, and CP solver augmentation with constrained text decoding~\citep{regin_et_al:LIPIcs.CP.2024.25}. 

Interactive decision aids such as \textsc{OptiChat} which is used to diagnose infeasible problems\citep{chen2025optichat,optichat}, \textsc{OptiGuide} used in supply chain \citep{optiguide}, and enterprise agents \citep{orderique2024domainadaptableprescriptiveai} let users iteratively query, diagnose, or adjust optimization runs. Meeting-scheduling assistants illustrate similar ideas for constraint programming~\citep{lawless2023meetmate}. 

\paragraph{Datasets and Evaluation:}
There exist attempts to create diverse, comprehensive benchmarks for the task of converting free-form natural language text to solver code. Optimization-centric corpora include \textsc{Nl4Opt}~\citep{lpwp}, \textsc{Nlp4Lp}~\citep{optimus}, \textsc{IndustryOR}~\citep{orlm}, \textsc{MILP} synthesis datasets~\citep{li2023synthesizingmixedintegerlinearprogramming}, Logic Grid Puzzles (LGP)~\citep{jabrayilzade2020lgpsolver}. As in our paper, there is continued interest in dataset curation. These include \textsc{Planetarium}~\citep{zuo2025planetariumrigorousbenchmarktranslating}, which  translates natural language descriptions into planning domain definition language (PDDL), \textsc{EHOP}~\citep{duchnowski2025ehopdataseteverydaynphard}, which is a collection of multiple versions of everyday NP-Hard problem and \textsc{DualSchool}~\citep{klamkin2025dualschoolreliablellmsoptimization}, which is aimed at evaluating the reliability of LLMs for optimization education. 

Evaluation metrics have evolved from exact string match, to evaluation and solut ion accuracy, to process-based scoring, exemplified by the \textsc{MAMO} benchmark \citep{mamo}. Specific instance data may also be abstracted from the generic problem statement. Meanwhile, \textsc{MOSDEX} proposes a format-agnostic exchange standard \citep{BloomSaltzmanKing2023MOSDEX}. 

To date, despite all these emerging literature, there exists no dataset for LLM Modelling copilots that enables benchmarking both satisfaction \emph{and} optimization problems, offers independence from underlying solvers and paradigms, blending realistic linguistic variety, and separating generic problem statements from specific instance data. The \textsc{Text2Zinc} dataset is designed to meet these criteria, thereby enabling a comprehensive assessment.

\section{Conclusion}
\label{sec:conclusion}
In this paper, we first formalized text-to-model translation problem and proposed several \textsc{Text2Model} copilots, along with an online leaderboard for serving the NLP \& OR communities. We then contributed the \textsc{Text2Zinc} dataset to enable benchmarking across both optimization and satisfaction problems in a solver-agnostic manner, along with an interactive editor with built-in AI assistant to speed-up data curation in future.


Our initial findings suggest that LLMs are not \textit{yet} a plug-and-play solution for combinatorial modeling despite their impressive capabilities in other domains. On several instances, the low solution accuracy across different strategies highlight the inherent challenges in translating natural language specifications into correct formal \textsc{MiniZinc} models. Nevertheless, the improved performance achieved through reasoning, grammar encoding, and agentic approaches offers promising directions. 

We emphasize that advancing modeling copilots depends on large-scale, high-quality datasets. We are grateful to the efforts of previous work that inspired and help shape our work. Our experiments with intermediate representations, particularly Knowledge Graphs, reveal that the path from natural language to executable and correct models is not straightforward. While such representations have potential, more research is needed to understand their best utilization. Future work should explore alternative representations, including named entities, semantic graphs, and agentic frameworks, which might better capture the nuances of formulating combinatorial problems.


\section*{Acknowledgments}
We would like to express our sincere gratitude to the creators of all the existing datasets for their hard work and contribution to the community.

\bibliographystyle{INFORMS/informs2014} 
\bibliography{references}

\newpage
\appendix
\section{Knowledge Graph Generation and Example}
\label{appendix:knowledge_graph_generation}
\subsection{Knowledge Graph Creation}
Given an optimization problem description and the nomenclature to be used for parameters.\newline
Please generate a knowledge-graph.\newline
\textbf{Problem Description}:\newline
\{problem\_description\}\newline
\textbf{Input Data Nomenclature}:\newline
\{data\_nomenclature\}\newline
\textbf{Steps to Generate a Knowledge Graph (KG)}:\newline
1. \textbf{Identify Parameters}:\newline
   - Extract each parameter individually, specifying its type and name.\newline
   - Ensure that parameter names are aligned with the predefined nomenclature.\newline
   - Record any explicit bounds for parameters, if specified.\newline
   - Validate against the problem description to ensure that all relevant parameters are included, checking for any implicitly stated ones.\newline
2. \textbf{Identify Variables}:\newline
   - Identify each variable, noting their types and names.\newline
   - Align variable names with the nomenclature provided, ensuring consistency.\newline
   - Determine and document any bounds (explicit or inferred) for the variables.\newline
   - Cross-verify with the problem description to confirm all necessary variables are accounted for, including those not explicitly mentioned in the nomenclature.\newline
3. \textbf{Identify Constraints}:\newline
   - Detail each constraint separately, recording its description, formula, and associated variables.\newline
   - Classify constraints based on their nature (e.g., linear, nonlinear, global etc..).\newline
   - Highlight if any constraints are global, impacting multiple variables or conditions across the model.\newline
4. \textbf{Identify Objective}:\newline
   - Clearly define the objective function, including its description, mathematical representation, and the variables it affects.\newline
   - Note the type of optimization (minimization, maximization, satisfy) and any relevant constraints tied to the objective.\newline
   - Ensure the objective aligns with the overall goal of the optimization problem as described.\newline
5. \textbf{Generate Turtle (TTL) Representation}:\newline
   - Construct the Turtle format representation using the gathered information on parameters, variables, constraints, and the objective.\newline
   - Sequentially organize the definitions of parameters, variables, constraints, and the objective within the Turtle file.\newline
   - Avoid generalizing the problem statement; focus on defining each component distinctly as outlined in the steps.\newline
Please do not generate any other token other than the knowledge graph itself. No titles, description or markup necessary.
\subsection{Example Knowledge Graph}
\label{appendix:knowledge_graph_example}
\textbf{Unstructured Description}\newline
A firm produces different goods using different raw materials. The firm has a limited amount of each raw material available. Each good requires a specific amount of raw materials to be produced. Each good results in an amount of revenue produced per unit. How much of each good should the firm produce to maximize its total revenue?\newline\newline
\textbf{Knowledge Graph}
\begin{lstlisting}[language=SPARQL, caption=Knowledge Graph in TTL format, basicstyle=\footnotesize\ttfamily,
    showspaces=false,
    showstringspaces=false,
    showtabs=false,
    keepspaces=true,
    breaklines=true,
    breakatwhitespace=true,
    columns=flexible]
@prefix : <http://example.org/firm_optimization#> .

# Parameters

:M a :Parameter ;
   :name "Number of different goods" ;
   :type "Integer" ;
   :description "Total number of different goods produced by the firm" .

:N a :Parameter ;
   :name "Number of different raw materials" ;
   :type "Integer" ;
   :description "Total number of different raw materials used by the firm" .

:Available a :Parameter ;
   :name "Available amount of raw material i" ;
   :type "1-D Array" ;
   :description "Amount of each raw material available for production" .

:Requirements a :Parameter ;
   :name "Requirement of raw material i to produce one unit of good j" ;
   :type "2-D Array" ;
   :description "Amount of each raw material required to produce one unit of each good" .

:Prices a :Parameter ;
   :name "Revenue from selling one unit of good j" ;
   :type "1-D Array" ;
   :description "Revenue earned by selling one unit of each good" .

# Variables

:UnitsProduced a :Variable ;
   :name "Units produced of good j" ;
   :type "1-D Array" ;
   :bounds "0 to infinity" ;
   :description "Number of units of each good produced" .

# Constraints

:MaterialAvailability a :Constraint ;
   :description "Each good's production must not exceed available raw materials" ;
   :formula "Sum(Requirements[i][j] * UnitsProduced[j] for j in 1..M) <= Available[i] for i in 1..N" .

# Objective

:MaximizeRevenue a :Objective ;
   :description "Maximize the total revenue from selling the goods" ;
   :formula "Sum(Prices[j] * UnitsProduced[j] for j in 1..M)" ;
   :type "Maximization" .
\end{lstlisting}
\section{\textsc{MiniZinc} Grammar Example}
\label{appendix:grammar_example}
Consider this simple grammar rule from \textsc{MiniZinc} that defines boolean literals:

\begin{lstlisting}
<bool-literal> ::= "false" | "true"
\end{lstlisting}

This rule states that a boolean literal can be either the word \texttt{false} or the word \texttt{true}. The vertical bar \texttt{|} means "or", and the quotes indicate exact text tokens.
When an LLM this constraint in the prompt, it understands and makes sure:
\begin{itemize}
\item Boolean values have exactly two valid forms
\item No other spellings (like \texttt{False}, \texttt{TRUE}, or \texttt{0/1}) are valid
\end{itemize}
This simple rule prevents the model from generating invalid boolean expressions like \texttt{maybe} or \texttt{yes}, ensuring syntactic correctness in the generated \textsc{MiniZinc} code.

\section{\textsc{MiniZinc} Code Generation Prompts}
\label{appendix:appendix_prompts}

We present the details of the prompts used in our experiments below.

\subsection{Baseline}
You are an expert MiniZinc developer.\newline
Generate Minizinc code from a given problem description with additional information about the parameters provided.\newline
The MiniZinc code should assume that the data needed, will be provided in a specific format through a .dzn file, so the generated code should assume the same names defined in the \textbf{input data}.\newline
Please do not generate any other token, except the MiniZinc code.\newline
\textbf{Problem Description}:\newline
\{problem\_description\}\newline
\textbf{Input Data Nomenclature}:\newline
\{data\_nomenclature\}
\subsection{Chain-of-Thought (CoT)}
You are an expert MiniZinc developer.\newline
Generate Minizinc code from a given problem description with additional information about the parameters provided.\newline
The MiniZinc code should assume that the data needed, will be provided in a specific format through a .dzn file, so the generated code should assume the same names, shapes and data-types defined in the \textbf{input data and examples}.\newline
When generating the code, follow this format:\newline
```\newline
\% Parameters\newline
\% Variables\newline
\% Constraints\newline
\% Objective\newline
```\newline
Also, make sure to follow the following principles when generating the code:\newline
\textbf{General Principles}:\newline
1. The generated code should assume that data will be provided via a ".dzn" file. Do not declare values directly from the input data nomenclature and examples within the MiniZinc model.\newline
2. Adhere to the input data nomenclature and examples precisely when declaring input parameter names and their data types.\newline
3. Use bounded variables whenever possible. If bounds are explicit (e.g., non-negative), include them as constraints.\newline
4. When defining arrays of variables, ensure bounds are integers. Apply element-wise constraints in a separate constraint block if bounds depend on array elements.\newline
5. When defining arrays of variables, ensure bounding constraints are applied separately rather than during initialization to avoid type mismatches.\newline
6. Define explicit bounds for all variables used in linear expressions, either in their declaration or through additional constraints.\newline
7. Separate constraints into distinct constraint blocks whenever possible.\newline
8. Use direct and succinct definitions for constraints in the model.\newline
9. Utilize global constraints as much as possible.\newline
10. Declare all parameters and sets before using them in other declarations to avoid circular dependencies and ordering issues.\newline
11. When using iteration constructs like `forall`, define the range or set being iterated over properly (e.g., use `1..M` instead of `M` for iteration).\newline
12. Ensure operands in operations are of compatible types to prevent coercion errors.\newline
13. Declare all identifiers (such as indices or ranges like `n`) before using them in any array or parameter declarations.\newline
14. Ensure type consistency in expressions to avoid coercion errors. Explicitly cast types if necessary.\newline
15. Ensure there is only one objective, which will be a maximization, minimization, or a satisfy problem. Do not forget the `solve` keyword.\newline
Please do not generate any other token, except the MiniZinc code.\newline
\textbf{Problem Description}:\newline
\{problem\_description\}\newline
\textbf{Input Data Nomenclature and Examples}:\newline
\{data\_nomenclature\}
\subsection{Knowledge Graph Code Generation}
You are an expert MiniZinc developer.\newline
Generate Minizinc code from using the following information:\newline
1. Problem Description: A formal description describing the optimization problem.\newline
2. Knowledge Graph: Detailing Parameters, Variables, Constraints and Objective.\newline
3. Input Data Nomenclature: The MiniZinc code should assume that the data needed, will be provided in a specific format through a .dzn file, so the generated code should assume the same names defined in the \textbf{input data}.\newline
Please do not generate any other token, except the MiniZinc code.\newline
\textbf{Problem Description}:\newline
\{problem\_description\}\newline
\textbf{Knowledge Graph}:\newline
```\newline
\{knowledge\_graph\}\newline
```\newline
\textbf{Input Data Nomenclature}:\newline
\{data\_nomenclature\}
\subsection{Code Validation}
You are an expert MiniZinc developer.\newline
The generated MiniZinc code failed to compile. Review and fix the code based on the error message, problem description, input parameters, and objective type.\newline
\textbf{Problem Description}:\newline
\{problem\_description\}\newline
\textbf{Input Data Nomenclature and Examples:}\newline 
\{data\_nomenclature\}\newline
\textbf{Objective Type}:\newline
\{objective\_type\}\newline
\textbf{Generated MiniZinc Code}:\newline
```minizinc\newline
\{minizinc\_code\}\newline
```\newline
\textbf{Error message after execution}:\newline
\{syntax\_error\_message\}\newline
\textbf{Validation Checklist}\newline
1. Fix the compilation error first - Address the specific issue indicated in the error message above.\newline
2. Ensure all parameters and variable names in `data.dzn` match the generated MiniZinc code.\newline
3. Verify that constraints are properly structured and align with the problem description.\newline
4. Check the objective function to confirm it is correctly set as:\newline
   - `minimize` if `\{objective\_type\}` is "minimization".\newline
   - `maximize` if `\{objective\_type\}` is "maximization".\newline
   - `satisfy` if `\{objective\_type\}` is "satisfaction".\newline
5. Ensure no syntax errors exist in the generated MiniZinc code.\newline
6. Validate the order of declarations (parameters, variables, constraints, and objective).\newline
7. Identify any missing components or inconsistencies.\newline
If any issues are found, revise the MiniZinc code accordingly. Output only the corrected MiniZinc code.
\subsection{Grammar Validation}
You are given a MiniZinc model that has syntax errors. Your task is to fix these syntax errors using the MiniZinc grammar specification provided below.\newline
\textbf{Problem Description}:\newline
\{problem\_description\}\newline
\textbf{Data Nomenclature}:\newline
\{data\_nomenclature\}\newline
\textbf{Current Code with Syntax Errors}:\newline
\{current\_code\}\newline
\textbf{Error message after execution}:\newline
\{syntax\_error\_message\}\newline
\textbf{MiniZinc Grammar Specification}:\newline
\{minizinc\_grammar\}\newline
Using the grammar specification above, carefully analyze the syntax errors and correct the code. Please provide the corrected MiniZinc code that adheres to the grammar specification and resolves all syntax errors. Output only the complete and corrected MiniZinc code without additional comments.
\subsection{Compositional}
\subsubsection{Parameters and Variables}\hfill\break
You are an expert MiniZinc developer.\newline
Generate MiniZinc code for the Parameters and Variables from a given problem description with additional information about input data provided.\newline
The MiniZinc code should assume that the data needed will be provided in a specific format through a .dzn file, so the generated code should assume the same names/data-types defined in the \textbf{input data and examples}.\newline
When generating the code, follow this format:\newline
```minizinc\newline
\% Parameters\newline
\% Variables\newline
```\newline
Also, make sure to follow the following principles when generating the code:\newline
\textbf{General Principles}:\newline
1. The generated code should assume that data will be provided via a ".dzn" file. Do not declare values directly from the input data nomenclature and examples within the MiniZinc model.\newline
2. Adhere to the input data nomenclature and examples precisely when declaring input parameter names and their data types.\newline
3. Use bounded variables whenever possible. If bounds are explicit (e.g., non-negative), include them as constraints.\newline
4. When defining arrays of variables, ensure bounds are integers. Apply element-wise constraints in a separate constraint block if bounds depend on array elements.\newline
5. When defining arrays of variables, ensure bounding constraints are applied separately rather than during initialization to avoid type mismatches.\newline
6. Define explicit bounds for all variables used in linear expressions, either in their declaration or through additional constraints.\newline
7. Declare all parameters and sets before using them in other declarations to avoid circular dependencies and ordering issues.\newline
8. Declare all identifiers (such as indices or ranges like `n`) before using them in any array or parameter declarations.\newline
9. Ensure that all indices and sets used in parameter and variable declarations are declared beforehand.\newline
10. When declaring variables, ensure they are appropriately typed (e.g., `int`, `float`).\newline
11. Variables should have meaningful names related to the problem description.\newline
12. Include comments to briefly describe each parameter and variable for clarity.\newline
\textbf{Problem Description}:\newline
\{problem\_description\}\newline
\textbf{Input Data Nomenclature and Examples}:\newline
\{data\_nomenclature\}
\subsubsection{Constraints}\hfill\break
You are an expert MiniZinc developer.\newline
Generate MiniZinc code for the Constraints from a given problem description with additional information about the parameters provided.\newline
Given the Parameters and Variables part of the code, generate only the constraints.\newline
When generating the code, follow this format:\newline
```minizinc\newline
\% Constraints\newline
```\newline
Also, make sure to follow the following principles when generating the code:\newline
\textbf{General Principles}:\newline
1. Separate constraints into distinct constraint blocks whenever possible.\newline
2. Utilize global constraints as much as possible.\newline
3. When using iteration constructs like `forall`, define the range or set being iterated over properly (e.g., use `1..M` instead of `M` for iteration).\newline
4. Ensure operands in operations are of compatible types to prevent coercion errors.\newline
5. Ensure type consistency in expressions to avoid coercion errors.\newline
6. Clearly comment on the purpose of each constraint for clarity and maintenance.\newline
7. Avoid hardcoding values; use parameters and variables instead.\newline
8. Use meaningful names for all constraint blocks.\newline
9. Only generate constraints and do not generate objective.\newline
\textbf{Problem Description}:\newline
\{problem\_description\}\newline
\textbf{Input Data Nomenclature and Examples}:\newline
\{data\_nomenclature\}\newline
\textbf{Parameters and Variables}:\newline
```minizinc\newline
\{parameters\_and\_variables\}\newline
```
\subsubsection{Objective}\hfill\break
You are an expert MiniZinc developer.\newline
Generate MiniZinc code for the Objective from a given problem description with additional information about the parameters, variables and constraints provided.\newline
Given the Parameters, Variables, and Constraints sections of the code, generate only the objective.\newline
When generating the code, follow this format:\newline
```minizinc\newline
\% Objective\newline
```\newline
Also, make sure to follow the following principles when generating the code:\newline
\textbf{General Principles}:\newline
1. Ensure there is only one objective, which will be a maximization, minimization, or a satisfy problem. Do not forget the `solve` keyword.\newline
2. Ensure the objective function aligns with the problem description.\newline
3. Verify the correct usage of variables, parameters and constraints in the objective function.\newline
\textbf{Problem Description}:\newline
\{problem\_description\}\newline
\textbf{Input Data Nomenclature and Examples}:\newline
\{data\_nomenclature\}\newline
\textbf{Parameters and Variables}:\newline
```minizinc\newline
\{parameters\_and\_variables\}\newline
```\newline
\textbf{Constraints}:\newline
```minizinc\newline
\{constraints\}\newline
```
\subsubsection{Stitch}\hfill\break
You are an expert MiniZinc developer.\newline
Given the Parameters, Variables, Constraints, and Objective sections of the code, stitch them into a complete solution for the optimization problem.\newline
When stitching the code, follow this format:\newline
```minizinc\newline
\% Parameters\newline
\% Variables\newline
\% Constraints\newline
\% Objective\newline
```\newline
Ensure the following principles for syntactic accuracy and logical consistency:\newline
\textbf{General Principles}:\newline
1. Verify that all intermediate sections (parameters, variables, constraints, objective) are consistent and correctly referenced.\newline
2. Confirm that the final MiniZinc code is syntactically accurate and logically coherent.\newline
3. Ensure that the code sections are properly integrated, maintaining the prescribed format.\newline
4. Check for and resolve any circular dependencies or ordering issues in declarations.\newline
5. Check for and resolve any coercion issues.\newline
5. Validate type consistency across all expressions and declarations.\newline
6. Utilize clear and concise comments to describe each section and its components.\newline
7. Make sure global constraints are utilized where applicable to enhance model efficiency.\newline
8. Ensure only one objective is defined, using the `solve` keyword appropriately.\newline
\textbf{Problem Description}:\newline
\{problem\_description\}\newline
\textbf{Input Data Nomenclature and Examples}:\newline
\{data\_nomenclature\}\newline
\textbf{Parameters and Variables}:\newline
```minizinc\newline
\{parameters\_and\_variables\}\newline
```\newline
\textbf{Constraints}:\newline
```minizinc\newline
\{constraints\}\newline
```\newline
\textbf{Objective}:\newline
```minizinc\newline
\{objective\}\newline
```\newline
\textbf{Note}: If the \texttt{.dzn} file is empty, we append the following instruction to the prompt: \textit{``All data and parameters are already included in the problem description above. You must embed all data directly in the MiniZinc model -- do not expect external \texttt{.dzn} files or assume data will be provided separately.''} Additionally, to avoid confusion from irrelevant solver references in the problem description, we append: \textit{``Generate MiniZinc code ONLY. Do NOT generate CPOPT, COPT, or any other format even if the problem description mentions it.''}

\section{Text2Zinc Statistics}
\label{appendix:text2zinc_stats}

In the following, we detail the characteristics of each data source and our specific contributions. 




\begin{itemize}
    \item \textbf{NLP4LP: 65 Problems}~\citep{optimus} A collection of 65 LP, MILP, and MIP problems, originally represented in structured natural language. We converted the data from JSON to DZN format via a Python script and refined problem statements for clarity. Where included, we verified our generated code against the objective values provided in the dataset. 
    \item \textbf{ComplexOR: 7 Problems}~\citep{complexor} A diverse collection of 37 operations research problems sourced from academic papers, textbooks, and industry applications, spanning domains such as supply chain optimization, scheduling, and warehouse logistics. We sampled seven sampled representative problems from this dataset.
    \item \textbf{LPWP: 5 Problems}~\citep{lpwp} This is a collection 1101 elementary-level linear programming (LP) problems collected from the Nl4Opt Competition~\citep{pmlr-v220-ramamonjison23a}. For these problems, we extracted parameters from problem descriptions into data files and modified problem descriptions accordingly.
    \item \textbf{CSPLib: 11 Problems}~\citep{csplib} A comprehensive library of 95 test problems for constraint solvers, featuring problems across various domains, including bin-packing, combinatorial mathematics, and scheduling. We included 11 problems from this collection and enhanced their existing \textsc{MiniZinc} solutions with detailed comments. 
    \item \textbf{Hakank's Collection: 22 Problems}~\citep{hakank} An extensive collection of over 1000 \textsc{MiniZinc} models spanning combinatorial problems, puzzles, operations research, and global constraints. We extracted problem features and descriptions from the extensively documented model files to standardize 22 problems.
\end{itemize}

Table~\ref{tab:comparison} shows the distribution of problem instances modeled with LP, MIP, and CP across datasets while \textsc{Text2Zinc} serves cross the domains. The unverified instances are also included in the dataset to encourage exploratory research, benchmarking, and expansion in future, bringing the total instance count to 1,775.

\begin{table}[t]
    \centering
    \begin{tabular}{lccc}
        \toprule
        & \# LP & \# MIP & \# CP \\ 
        \midrule
        NLP4LP     &  54 &  13&  0\\
        ComplexOR  &  25 &  12&  0\\
        LPWP       &  1101 &  0&  0\\
        \textsc{Text2Zinc} &  64 & 31 & 15 \\
        \bottomrule
    \end{tabular}
    \caption{Distribution of problem types across datasets (only displaying verified Text2Zinc instances).}
    \label{tab:comparison}
\end{table}


Table 6 and 7 report the distribution of instances across source datasets and the distribution of objectives in these instances.

\begin{adjustbox}{minipage=0.48\textwidth,valign=t}
\centering
\captionof{table}{Distribution of Sources}
\begin{tabular}{>{\bfseries}l l}
\toprule
Source & \# of Problems \\
\midrule
ComplexOR & 27 \\
NLP4LP & 131 \\
Hakank & 393 \\
IndustryOR & 100 \\
MAMO & 863 \\
NL4OPT & 245 \\
CSPLib & 11 \\
LPWP & 5 \\
\bottomrule
\end{tabular}
\end{adjustbox}
\hfill
\begin{adjustbox}{minipage=0.48\textwidth,valign=t}
\centering
\captionof{table}{Distribution of Objectives}
\begin{tabular}{>{\bfseries}l l}
\toprule
Objective & \# of Problems \\
\midrule
Minimization \\\& Maximization & 136 \\
Maximization & 340 \\
Minimization & 1009 \\
Satisfaction & 242 \\
Unlabelled & 48 \\
\bottomrule
\end{tabular}
\end{adjustbox}
\newline\newline


\vspace{-0.2cm}
\section{\textsc{OptiMind} Results}
\label{appendix:optimind}

Table~\ref{tab:optimind} reports running our co‑pilots in an ``as‑is" manner by replacing GPT‑5.2 with \textsc{OptiMind} on the \textsc{Text2Zinc} dataset. We note that \textsc{OptiMind} is fine‑tuned specifically for generating \textsc{GurobiPy} models, and may therefore require re‑designing our co‑pilots and prompts to better align with \textsc{Gurobi}. Our goal in this paper is to remain paradigm‑ and solver‑agnostic, hence, we do not pursue such adaptations here. Interestingly, \textsc{Gala} (\cite{cai2025galagloballlmagents}), which also employs GPT‑OSS‑20b, reports competitive performance when generating \textsc{MiniZinc} models. Our findings suggest that \textsc{OptiMind}'s fine‑tuning GPT‑OSS‑20b to specialize on \textsc{Gurobi} may degrade its broader modeling capabilities. 

\begin{table}[b]
\centering
\caption{\textsc{OptiMind} results on \textsc{Text2Zinc}}
\label{tab:optimind}
\begin{adjustbox}{max width=\textwidth}
\begin{tabular}{@{}lc|cc|cc|cc|cc|cc|cc@{}}
\toprule
& & \multicolumn{2}{c|}{\textsc{NLP4LP}} & \multicolumn{2}{c|}{\textsc{ComplexOR}} & \multicolumn{2}{c|}{\textsc{LPWP}} & \multicolumn{2}{c|}{\textsc{CspLib}} & \multicolumn{2}{c|}{\textsc{Hakank}} & \multicolumn{2}{c}{Average} \\
& & \multicolumn{2}{c|}{(n=65)} & \multicolumn{2}{c|}{(n=7)} & \multicolumn{2}{c|}{(n=5)} & \multicolumn{2}{c|}{(n=11)} & \multicolumn{2}{c|}{(n=22)} & \multicolumn{2}{c}{(n=110)} \\
copilot Strategy & LLM Calls & $E_{acc}$ & $S_{acc}$ & $E_{acc}$ & $S_{acc}$ & $E_{acc}$ & $S_{acc}$ & $E_{acc}$ & $S_{acc}$ & $E_{acc}$ & $S_{acc}$ & $E_{acc}$ & $S_{acc}$ \\
\midrule
\textsc{Zero Shot} & 1 & 3.08 & 0.00 & 0.00 & 0.00 & 0.00 & 0.00 & \best{18.18} & \best{9.09} & 4.55 & 0.00 & 4.55 & 0.91 \\
\textsc{Chain-of-Thought (CoT)} & 1 & \best{7.69} & \best{3.08} & 0.00 & 0.00 & \best{20.00} & \best{20.00} & 9.09 & 0.00 & 4.55 & \best{4.55} & \best{7.27} & \best{3.64} \\
\bottomrule
\end{tabular}
\end{adjustbox}
\end{table}

\section{GPT-4 and GPT-4O Reasoning Results}
\label{appendix:gpt4_results}
\begin{table}[t]
\renewcommand{\arraystretch}{1}
\centering
\begin{tabular}{llccc}
    \toprule
    \textbf{copilot Strategy} & \textbf{LLM} & \textbf{\makecell[c]{Execution\\Accuracy (\%)}} & \textbf{\makecell[c]{Solution\\Accuracy (\%)}} & \textbf{\makecell[c]{\#LLM\\Query}} \\
    \midrule
    \multicolumn{5}{l}{\textit{Single-Call Strategies}} \\
    \quad Baseline & GPT-4 & 32.73 & 17.27 & 1 \\
    \addlinespace[0.5em]
    \multirow{2}{*}{\quad Chain-of-Thought (CoT)} 
    & GPT-4o & \underline{60.91} & \underline{34.55} & 1 \\
    & GPT-4 & 57.27 & 28.18 & 1 \\
    \addlinespace[0.5em]
    \multicolumn{5}{l}{\textit{Multi-Call Strategies}} \\
    \quad Knowledge Graph & GPT-4 & 48.18 & \underline{25.45} & 2 \\
    \addlinespace[0.5em]
    \multirow{2}{*}{\quad CoT + Code Validation} 
    & GPT-4o & \textbf{80.91} & \textbf{41.82} & 2 \\
    & GPT-4 & 57.27 & 28.18 & 2 \\
    \addlinespace[0.5em]
    \quad CoT + Grammar Validation & GPT-4 & \underline{62.73} & 22.73 & 2 \\
    \addlinespace[0.5em]
    \multirow{2}{*}{\quad CoT + Code \& Grammar Validation} 
    & GPT-4o & \underline{73.64} & \underline{40.00} & 3 \\
    & GPT-4 & 70.00 & 24.55 & 3 \\
    \addlinespace[0.5em]
    \multicolumn{5}{l}{\textit{Agentic Strategies}} \\
    \quad Agentic & GPT-4 & \underline{43.64} & 20.00 & 4 \\
    \addlinespace[0.5em]
    \quad Agentic + Code Validation & GPT-4 & \underline{43.64} & \underline{20.91} & 5 \\
    \bottomrule
\end{tabular}
\caption{Experimental results with GPT-4 and GPT-4o comparing execution and solution accuracy across different prompting strategies and models on the \textsc{Text2Zinc} dataset. \underline{Underlined} indicates best performance within each strategy and \textbf{Bold} indicates the overall best performance. We also report the number of LLM queries required.}
\label{tab:experimental-results}
\end{table}
\subsection{Single-Call Strategies}
Our analysis of single-call approaches reveals several important insights:

\medskip
\noindent \textbf{Baseline:} The baseline approach with GPT-4 achieves 32.73\% execution accuracy and 17.27\% solution accuracy, which serves as the lower bound performance. These results highlight that merely exposing the LLM to problem descriptions and input data is insufficient for complex constraint modeling tasks.

\medskip
\noindent \textbf{Chain-of-Thought (CoT):} CoT dramatically improves performance across both models, with GPT-4o achieving 60.91\% execution accuracy and 34.55\% solution accuracy, while GPT-4 reaches 57.27\% and 28.18\% respectively. This substantial improvement demonstrates that structured step-by-step reasoning is beneficial for generating constraints models, nearly doubling the baseline.

\subsection{Multi-Call Strategies}

The multi-call approaches reveal subtle trade-offs between computational cost and performance:

\medskip
\noindent \textbf{Knowledge Graph:} The KG approach achieves 48.18\% execution accuracy and 25.45\% solution accuracy with GPT-4, representing a clear improvement over baseline. However, whether this improvement justifies the additional API call for creating potentially noisy knowledge graphs remains questionable. More sophisticated prompt engineering for KG creation might yield better results and warrants future experimentation; however, given the modest improvements, we opted to focus our GPT-4o experiments on more promising strategies.

\medskip
\noindent \textbf{CoT + Code validation:} CoT combined with code validation achieves the highest overall performance, with GPT-4o reaching 80.91\% execution accuracy and 41.82\% solution accuracy. Notably, this strategy shows much greater improvement for GPT-4o than for GPT-4, suggesting that reasoning models might be better positioned to leverage validation feedback effectively.

\medskip
\noindent \textbf{CoT + Grammar validation:} CoT with grammar validation shows positive impact particularly for GPT-4 (62.73\% execution accuracy), though the solution accuracy improvement is modest (22.73\%). While grammar guidance clearly helps with syntactic correctness, directly attaching grammar files to prompts may not be the best approach. Alternative methods for leveraging grammatical rules deserve further exploration.

\medskip
\noindent \textbf{CoT + Code \& Grammar Validation:} The combination of both code and grammar validation produces decent improvements for both models, but surprisingly does not exceed code validation alone. For GPT-4o, the combined approach achieves 73.64\% execution accuracy and 40.00\% solution accuracy, slightly lower than code validation alone. This counter-intuitive result reinforces our earlier observation that direct grammar file incorporation through prompting may not be the best combination. 

\subsection{Agentic Strategies}

The agentic frameworks reveal important limitations of task decomposition:

\medskip
\noindent \textbf{Agents w/ and w/o code validation:} Surprisingly, agentic strategies show modest performance with GPT-4 achieving 43.64\% execution accuracy and 20.00\% solution accuracy. Even with code validation added, the improvement is minimal (20.91\% solution accuracy). Simply increasing the number of LLM calls decomposed into simpler agents do not necessarily guarantee better outcomes.

\subsection{General Observations}

When considering general observations across all approaches, the following patterns emerge:

\medskip
\noindent \textbf{Chain-of-Thought is a strong method:} A simple CoT approach with a reasoning model already delivers substantial metric improvements, serving as an essential competitor for more complex strategies.

\medskip
\noindent \textbf{Reasoning Model superiority:} GPT-4o consistently outperforms GPT-4 across all comparable strategies, reinforcing its advantages for complex reasoning tasks. This performance gap is particularly pronounced in validation-enhanced approaches.

\medskip
\noindent \textbf{Domain-specific guidance shows promise:} Both code and grammar validation demonstrate that domain-specific additional guidance can improve metrics for both models, though the optimal integration strategies require further investigation.

\medskip
\noindent \textbf{Over-segmentation may harm:} The relatively poor performance of compositional approaches suggests that excessive task decomposition negatively impacts overall quality of the modal code. The challenge is the integration between intermediate outputs that outweigh the benefits of specialized sub-task handling.

\medskip
\noindent \textbf{Failure analysis:} In our observation, syntax errors are the primary cause of execution failures. This can be attributed to the LLM's limited training on \textsc{MiniZinc}'s specialized syntax and constraints, which differ significantly from widely-used programming languages such as Python. More concretely, in Appendix~\ref{appendix:error_analysis}, we document the detailed breakdown of these errors for reference.

\medskip
\noindent \textbf{Execution-Solution accuracy gap persists:} Consistently lower solution accuracies across different strategies continue to highlight the complexity of generating semantically correct models. This gap suggests that while models can produce syntactically valid code, capturing the underlying optimization logic is not immediate. Our call-to-action to the community is to close this performance gap, and our open-source suite is designed to facilitate this goal.

\newpage
\section{Error Analysis}
\label{appendix:error_analysis}
\begin{table}[H]
\centering
\renewcommand{\arraystretch}{1.0}
{\footnotesize
\begin{tabular}{l p{10cm}}
\hline
\textbf{Error Type} & \textbf{Commentary} \\
\hline
Syntax Errors & These errors indicate issues with the \textsc{MiniZinc} code syntax. They may include unexpected tokens, missing semicolons, or incorrect use of language constructs. For example: \texttt{Error: syntax error, unexpected where, expecting ')'}. \\
\hline
Undefined Identifiers & These errors are due to the use of variables or identifiers that have not been declared or are out of scope. Example: \texttt{Error: type error: undefined identifier `i', did you mean `X'?}. \\
\hline
Array and Indexing Issues & Errors related to improper use of arrays or indexing problems. This can include out-of-bounds errors or incorrect array dimensions. Example: \texttt{Error: type error: initialisation value for `Downtime' has invalid type-inst: expected `array[int] of int', actual `array[int,int] of int'}. \\
\hline
Function Not Found Errors & These errors occur when attempting to use functions or predicates that do not exist in the current \textsc{MiniZinc} environment or have incorrect signatures. This can happen with missing imports or incorrect function names. Example: \texttt{Error: type error: no function or predicate with this signature found: `..o(float)'}. \\
\hline
Variable Redefinition Errors & These errors happen when the same variable is assigned or defined multiple times inappropriately, violating \textsc{MiniZinc}'s variable declaration rules. Example: \texttt{Error: type error: multiple assignment to the same variable}. \\
\hline
Flattening Errors & These errors occur during the model flattening phase when \textsc{MiniZinc} attempts to convert the high-level model into a lower-level representation. Often related to unbounded variables or overly complex expressions that cannot be properly linearized. Example: \texttt{Error: flattening error: unbounded coefficient in linear expression}. \\
\hline
Timeout Errors & These errors occur when the solver execution exceeds the specified time limit, typically indicating computational complexity issues or problems that are too difficult to solve within the given timeframe of 60 seconds. \\
\hline
Solver Limitation Errors & These errors arise when the selected solver cannot handle certain types of mathematical formulations or constraints. This includes cases where quadratic or non-linear constraints are required but not supported by the solver. Example: \texttt{Unable to create linear formulation for quadratic constraint}. \\
\hline
Missing Data in DZN Files & These errors occur when the necessary data is not provided in the DZN files. This may lead to missing parameters or sets that are required by the model. Example: \texttt{Error: type error: variable `K' must be defined (did you forget to specify a data file?)}. \\
\hline
\end{tabular}}
\vspace{0.2cm}
\caption{Categorization of \textsc{MiniZinc} Errors}
\end{table}

\end{document}